\documentclass{article}

\usepackage[utf8]{inputenc}%
\usepackage[T1]{fontenc}%
\usepackage[hyperfootnotes=false]{hyperref}%
\usepackage{url}%
\usepackage{booktabs}%
\usepackage{amsfonts}%
\usepackage{nicefrac}%
\usepackage{microtype}%
\usepackage{xcolor}%

\input{packages}

\PassOptionsToPackage{numbers, compress}{natbib}
\usepackage[main, final]{neurips_2026}

\definecolor{shade}{RGB}{220, 230, 255}
\definecolor{optimaTeal}{RGB}{0,128,128}
\definecolor{TaskNameColor}{RGB}{8,8,184}
\newcommand{\taskname}[1]{\textcolor{TaskNameColor}{#1}}
\newcommand{\avgname}[1]{\textcolor{optimaTeal}{#1}}
\newcommand{\oursrow}{\rowcolor{shade}}
\newcommand{\ourssep}{\arrayrulecolor{white}\specialrule{1pt}{0pt}{0pt}\arrayrulecolor{black}}
\colorlet{PatchShade}{shade}

\definecolor{AlgKeywordColor}{RGB}{0, 102, 153}
\definecolor{AlgVarColor}{RGB}{128, 0, 128}
\definecolor{AlgCommentColor}{RGB}{0, 128, 0}
\algrenewcommand{\algorithmicrequire}{\textbf{\color{AlgKeywordColor}Input:}}
\algrenewcommand{\algorithmicensure}{\textbf{\color{AlgKeywordColor}Output:}}
\algrenewcommand{\algorithmiccomment}[1]{\hfill{\color{AlgCommentColor}\everymath{\color{AlgCommentColor}}$\triangleright$ #1}}

\newif\ifshowtodos
\showtodostrue

\newif\ifshowrev
\showrevfalse%
\newcommand{\rev}[1]{\ifshowrev\textcolor{blue}{#1}\else#1\fi}
\newcommand{\revcolor}{\ifshowrev\color{blue}\fi}
\newenvironment{revblock}{\revcolor}{}

\newcommand{\oursname}{\textsc{PATCH}}
\newcommand{\ours}{\texorpdfstring{\oursname}{PATCH}\xspace}
\newcommand{\our}{\texorpdfstring{\oursname}{PATCH}}
\newcommand{\ourstile}{\texorpdfstring{\oursname\textsuperscript{Tile}}{PATCH-Tile}\xspace}
\newcommand{\oursjoint}{\texorpdfstring{\oursname\textsuperscript{Joint}}{PATCH-Joint}\xspace}
\definecolor{modelblue}{HTML}{646464}

\definecolor{KeywordColor}{RGB}{0, 102, 153}%
\definecolor{VarColor}{RGB}{128, 0, 128}%
\definecolor{CommentColor}{RGB}{0, 128, 0}%

\def\mM{{\bm{M}}}

\def\mP{{\bm{P}}}

\def\mS{{\bm{S}}}

\def\mW{{\bm{W}}}

\title{\our: Learnable Tile-Level Hybrid Sparsity for LLMs}

\author{%
  Younes Hourri\thanks{Equal contribution.} \\
  Department of Computer Science \\
  University of Toronto, Canada \\
  \texttt{younes@cs.toronto.edu} \\
  \And
  Mohammad Mozaffari\footnotemark[1] \\
  Department of Computer Science \\
  University of Toronto, Canada \\
  \texttt{mmozaffari@cs.toronto.edu} \\
  \And
  Maryam Mehri Dehnavi \\
  Department of Computer Science \\
  University of Toronto, Canada \\
  \texttt{mmehride@cs.toronto.edu}
}

\begin{document}

\maketitle
\setcounter{footnote}{0}

\begin{abstract}
Large language models (LLMs) deliver impressive performance but incur prohibitive memory and compute costs at deployment. Model pruning is an effective way to reduce these overheads, yet existing approaches face challenges: unstructured sparsity, where nonzeros can appear anywhere, preserves accuracy but yields irregular access patterns that prevent GPU acceleration, while semi-structured 2:4 sparsity is hardware-friendly but enforces a rigid 50\% pattern that degrades model quality. To bridge this gap, we introduce \ours, a hybrid sparsity framework that enables a continuous sparsity ratio between 0\% and 50\%. \ours partitions weight matrices into tiles, assigning each tile to be either dense or 2:4 sparse via a learnable mask selection mechanism. This design provides fine-grained control over accuracy–acceleration tradeoffs and supports non-uniform sparsity across layers, leading to superior overall quality. Across models from 0.5B to 13B parameters, \ours consistently narrows the gap to dense accuracy while delivering practical speedups. For instance, on LLaMA-2 7B with an A6000 GPU, \ours achieves 1.18×–1.38× end-to-end speedup over dense baselines while improving accuracy by 0.37\%–2.96\% compared to the state-of-the-art 2:4 pruning method, MaskLLM.%
\footnote{\rev{Code is available at \url{https://github.com/Paramathic/patch}, and the pruned checkpoints are released at \url{https://huggingface.co/collections/mohammad-mozaffari/patch}.}}
\end{abstract}

\section{Introduction}

Recent advancements in large language models (LLMs) have revolutionized natural language processing, enabling breakthroughs in understanding and generating human language ~\citep{gemini2.5, llama4}. These models power diverse applications, such as conversational agents and automated content creation ~\citep{llm_app1, llm_app2}. However, their extensive parameter counts, often in the billions, result in significant memory overhead and high inference costs ~\citep{jsq, affinequant}. This computational burden has driven the need for efficient model compression techniques.

\rev{Model compression rests on three main pillars that have matured unevenly: sparsity~\citep{obs, obd}, quantization~\citep{quarot, qtip}, and low-rank approximation~\citep{qera, caldera}, a triad referred to as the \emph{Compression Trinity}~\citep{trinity_hub, trinity_thesis}. Quantization reduces the precision of model parameters and now compresses LLMs effectively while preserving performance, and the pillars compose, as in joint sparse, quantized, and low-rank compression~\citep{slim}. Sparsity, which lowers memory and computational demands by setting many parameters to zero, lags behind: it still struggles to maintain model accuracy while delivering practical speedups, and closing this gap is the subject of this paper.}

\begin{figure}[t]
\centering
\includegraphics[width=\linewidth]{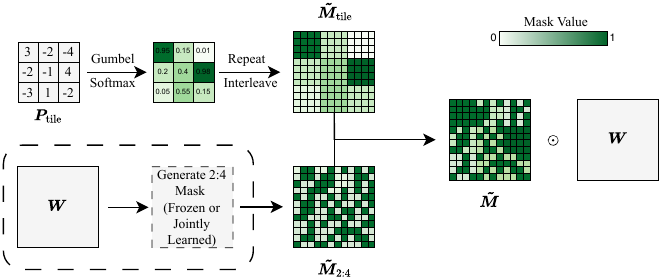}
\caption{Illustration of the PATCH learning process for generating tile-level hybrid masks. Each tile is parameterized by a learnable distribution and sampled with Gumbel Softmax to produce $\Tilde{\mM}_{\text{tile}}$. The dense probability is expanded and merged with a 2:4 mask $\Tilde{\mM}_{2:4}$, which can be fixed or jointly learned during training, yielding $\Tilde{\mM}$. The final mask assigns each tile to remain dense or follow the 2:4 pattern, enabling flexible sparsity across the weight matrix.}
\label{fig:overview}
\end{figure}

Unstructured sparsity, which permits non-zero elements to appear anywhere in the matrix, can match dense model accuracy due to its flexibility in sparsity allocation~\citep{wanda, sparsegpt, cerebras_sparsity}. However, its irregular memory access patterns hinder acceleration on modern hardware like GPUs~\citep{flashllm, spinfer}. As a result, unstructured sparsity fails to deliver practical speedups, motivating the search for more hardware-friendly sparsity techniques.

Semi-structured sparsity patterns, such as the 2:4 pattern~\citep{2:4} supported by NVIDIA and AMD GPUs, provide practical speedups in large-scale model inference. However, unlike unstructured sparsity, which offers greater flexibility, 2:4 enforces rigid rules by requiring at least two of every four consecutive elements to be zero. This rigidity often leads to significant accuracy loss when models are pruned using one-shot methods~\citep{wanda, sparsegpt, thanos, proxsparse}. MaskLLM~\citep{maskllm} mitigates this issue by learning sparsity masks end-to-end, but pruned models still lag behind their dense counterparts in accuracy. Moreover, recent studies show that sparsity should be allocated non-uniformly (adaptively) across layers for optimal performance~\citep{owl, rethinkingvaluetransformercomponents, layeradaptivesparsitymagnitudebasedpruning}, whereas 2:4 sparsity enforces a fixed, uniform allocation. These limitations indicate that relying solely on 2:4 sparsity is insufficient, underscoring the need for hybrid approaches.\footnote{\rev{Additional related work is discussed in Appendix~\ref{sec:related_work}.}}

To address the challenges of LLM pruning, while providing accelerated inference, we propose \textbf{P}runing with a Le\textbf{a}rnable \textbf{T}ile-level \textbf{C}onfiguration for \textbf{H}ybrid Sparsity (\textbf{\our}). \ours learns a hybrid mask that partitions each weight matrix into hardware-friendly tiles, designating each tile as either dense (0\% sparsity) or 2:4 sparse (50\% sparsity). This adaptive mask allows the matrix to realize an effective global sparsity ratio anywhere between 0\% and 50\%, balancing accuracy in critical regions with hardware-friendly sparsity elsewhere. This design unites the hardware acceleration benefits of 2:4 sparsity with the flexibility of unstructured allocation, allowing sparsity to adapt to the varying importance of different layers. By jointly optimizing the sparsity within 2:4 tiles and the tile-level patterns during training, \ours achieves higher accuracy than uniform sparsity across layers. Moreover, for resource-constrained settings, we offer a variant of \ours that tunes only the dense tiles while freezing the initial 2:4 mask. Importantly, \ours is compatible with tile-level sparsity acceleration libraries and compilers such as STOICC~\citep{stoicc}, making it the first hybrid sparsity method to demonstrate practical speedups. For example, on LLaMA-2 7B running on a consumer-grade A6000 GPU, \ours achieves 1.18$\times$–1.38$\times$ end-to-end speedup over the dense baseline while improving accuracy by 0.37\%–2.96\% compared to the state-of-the-art 2:4 pruning method, MaskLLM.

\section{Preliminaries}

\paragraph{Differentiable sampling.} Sampling from a categorical distribution is inherently non-differentiable, which poses challenges for gradient-based optimization. The Gumbel Softmax~\citep{gumbelsoftmax} addresses this by combining the Gumbel-Max reparameterization trick together with a softmax relaxation. The reparameterization expresses the sampling process by decoupling the deterministic log-probabilities $p \in \mathbb{R}^n$ from the stochastic perturbations $z \in \mathbb{R}^n$ introduced by Gumbel noise, which emulate random draws from the distribution. The subsequent softmax yields a differentiable approximation to categorical sampling:

\begin{equation}
    \mathrm{GS}(p;\, \tau, \kappa)_k =  \frac{\exp((\kappa p_k + z_k) / \tau)}{\sum_j \exp((\kappa p_j + z_j) / \tau)}
    \label{eq:gumbel_softmax}
\end{equation}

where $z_k = -\log(-\log(u_k))$ with $u_k \sim \mathrm{Uniform}(0,1)$. The resulting vector $\mathrm{GS}(p;\, \tau, \kappa) \in \mathbb{R}^n$ is a soft index vector whose entries $\mathrm{GS}(p;\, \tau, \kappa)_k$ represent the relaxed probability of selecting class $k$.

Additionally, the temperature parameter $\tau$ controls the hardness of the sampled index. Lower values of $\tau$ yield a more peaked distribution, causing $\mathrm{GS}(p;\, \tau, \kappa)$ to converge to a one-hot vector as $\tau \rightarrow 0$. 

\paragraph{Learnable 2:4 mask.} MaskLLM~\citep{maskllm} formulates 2:4 mask selection as a learnable probabilistic process over the six possible patterns. The underlying weights remain fixed, while training shifts the categorical distribution to favor masks that preserve better pruning performance. The mask for each four consecutive elements can be parameterized with a vector $p \in \mathbb{R}^{6 \times 1}$. Scaling this vector to a weight matrix  $\mW \in \mathbb{R}^{d_1 \times d_2}$ will result in $\mP_{\text{2:4}} \in \mathbb{R}^{\frac{d_1 d_2}{4} \times 6}$ as the mask search parameters. The resulting mask can be computed as in Equation~\ref{eq:2_4_mask}, where $\Tilde{\mM}_{\text{2:4}} \in [0,1]^{d_1 \times d_2}$ denotes the 2:4 soft mask, obtained as a weighted average over the candidate masks, and $\mS \in \mathbb{R}^{6 \times 4}$ is the matrix containing these six candidates as its rows.\footnote{We will refer to a mask value of $1$ as \textit{keeping} the corresponding weight and a value of $0$ as \textit{pruning} it.}

\begin{equation}
    \Tilde\mM_{\text{2:4}} =  \text{\tt {reshape}}(\mathrm{GS}(\mP_{\text{2:4}}; \tau, \kappa) \times \mS,  \mathbb{R}^{d_1 \times d_2})
    \label{eq:2_4_mask}
\end{equation}

The scaling factor $\kappa$ in Equation~\ref{eq:gumbel_softmax} multiplies the logits $p$ before adding the Gumbel noise $z$, thereby controlling their relative influence. Small $\kappa$ values let the noise dominate, encouraging exploration across candidate masks, while larger $\kappa$ values amplify the logits and make the sampling more deterministic.

\section{\our}
\label{sec:patch}

To overcome the rigidity of fixed 50\% 2:4 sparsity, we introduce \ours. \ours learns a structured mask, optimized on top of frozen weights, that is partitioned into tiles, where each tile decides whether its corresponding weights remain dense or are pruned with a 2:4 pattern. This design preserves accuracy in sensitive regions while exploiting hardware-accelerated sparsity elsewhere. Unlike fixed 2:4 sparsity, which enforces the same pattern across all weights, \ours adapts at the tile level by assigning dense tiles to critical regions and sparse tiles elsewhere.

Finding the optimal allocation of dense tiles (value 1) and sparse tiles (2:4 pattern) within a mask is a combinatorially difficult problem, as the number of possible configurations grows rapidly with the number of tiles across the LLM. By also modelling this problem as a probabilistic sampling process, and adjusting the probability of each tile (and the 2:4 patterns within sparse tiles), \ours can efficiently explore the space of configurations and converge toward masks that balance accuracy and sparsity. The mask distributions are learned end-to-end by training the Gumbel–Softmax logits while keeping the model weights frozen. We address this challenge by formulating mask selection as two coupled subproblems: \textit{(1) selecting which tiles are dense or sparse}, and \textit{(2) choosing the 2:4 sparsity pattern within sparse tiles}.

\paragraph{Tile-based pruning of LLMs.} We associate each parameter matrix $\mW \in \mathbb{R}^{d_1 \times d_2}$ with a grid of tile-level distributions, each parameterized by a learnable logit. Collectively, these form $\mP_{\text{tile}} \in \mathbb{R}^{\tfrac{d_1}{b_1} \times \tfrac{d_2}{b_2}}$, where each entry specifies the unnormalized score of keeping the corresponding $b_1 \times b_2$ tile fully dense. To create a two-class distribution (keep dense vs.\ prune), we concatenate a fixed zero to each logit, yielding $[\mP_{\text{tile}}, 0] \in \mathbb{R}^{\tfrac{d_1}{b_1} \times \tfrac{d_2}{b_2} \times 2}$. After applying Gumbel–Softmax, we broadcast the dense probabilities across their respective $b_1 \times b_2$ region (since the weighted average of the two outcomes reduces to $p_{\text{dense}} \cdot 1 + p_{\text{prune}} \cdot 0 = p_{\text{dense}}$), so that all elements of a tile receive the same mask value. Formally,

\begin{equation}
    \Tilde{\mM}_{\text{tile}} = \mathrm{GS}([\mP_{\text{tile}},0]; \tau, \kappa)_{:,:,0} \otimes \mathbf{1}.
    \label{eq:tiled_mask}
\end{equation}

This yields the tile-level mask $\Tilde{\mM}_{\text{tile}} \in [0,1]^{d_1 \times d_2}$ in Equation~\ref{eq:tiled_mask}, where $\mathbf{1} \in \mathbb{R}^{b_1 \times b_2}$ is an all-ones matrix and $\otimes$ denotes the Kronecker product.

\paragraph{Joint optimization with sparse mask.} To fully determine the effective sparsity pattern, the tile-level mask must be combined with the fine-grained 2:4 mask. Assuming that the 2:4 mask $\Tilde{\mM}_{2:4}$ is generated using Equation~\ref{eq:2_4_mask}, \ours combines it with the tile mask $\Tilde{\mM}_{\text{tile}}$ as shown in Equation~\ref{eq:tile_merge}. The resulting soft mask interpolates between dense and sparse behavior: values of $\Tilde{\mM}_{\text{tile}}$ close to one make the tile predominantly dense, while values close to zero shift the tile toward the soft 2:4 mask pattern defined by $\Tilde{\mM}_{\text{2:4}}$. Thus, $\Tilde{\mM}$ can be understood as a per-tile weighted average of the dense option and the 2:4 patterns, with $\Tilde{\mM}_{\text{tile}}$ determining the relative contribution of each. An overview of the process is provided in Figure~\ref{fig:overview}.

\begin{equation}
    \Tilde{\mM} = \Tilde{\mM}_{\text{tile}} + \left(1 - \Tilde{\mM}_{\text{tile}}\right) \odot \Tilde{\mM}_{\text{2:4}}
    \label{eq:tile_merge}
\end{equation}

\paragraph{Learning masks with targeted sparsity.} \ours uses a novel regularization term to achieve a flexible 0\%–50\% sparsity ratio across the model by controlling the number of dense tiles. Unlike traditional regularization methods like weight decay, which produce non-deterministic sparsity ratios, our term penalizes deviations from the target sparsity, enabling precise control. This global sparsity approach prunes sensitive linear layers less aggressively while setting redundant weight elements to zero, offering greater flexibility than fixed per-layer sparsity. We directly compare global versus per-layer sparsity regularization in $\S$ \ref{sec:results}.

\paragraph{Training objective.}
The overall training objective, as shown in Equation \ref{eq::loss}, of \ours combines three components: the standard modeling loss, 
a sparsity regularization term that enforces the target density of the model $\rho$,  and a weight regularization term (as in MaskLLM) that promotes larger weight magnitudes and gradient propagation.  
Formally,
\begin{equation}
    \mathcal{L} 
    = \mathcal{L}_{LM}\!\left(x ; \{\Tilde{\mM}_i \odot \mW_i\}_i\right) 
    + \lambda_1 \left\lvert \frac{\sum_i \lVert \Tilde{\mM}_i \rVert_1}{\sum_i \lVert \mW_i \rVert_0} - \rho \right\rvert 
    - \lambda_2 \frac{\sum_i \lVert \Tilde{\mM}_i \odot \mW_i \rVert_2^2}{\sum_i \lVert \mW_i \rVert_2^2}
    \label{eq::loss}
\end{equation}
\rev{where $i$ indexes the prunable weight matrices of the model; since mask entries lie in $[0,1]$, $\lVert \Tilde{\mM}_i \rVert_1$ is the (soft) number of kept weights.}

Following MaskLLM, we progressively \textit{decrease} $\tau$ and \textit{increase} $\kappa$ during training so that the Gumbel-Softmax distribution converges to a clear one-hot choice of mask by the end of training.

\paragraph{Inference.}  After training, the sign of each logit in $\mP_{\text{tile}}$ determines the final mask. Since a zero logit is concatenated to represent the sparse class (Equation~\ref{eq:tiled_mask}), positive values correspond to the dense option, while negative values correspond to the sparse option. The complete procedure is outlined in Algorithm~\ref{alg:patch}.

\begin{algorithm}[t]
\caption{Joint Tile \& 2{:}4 Mask Learning}
\label{alg:patch}
\begin{algorithmic}[1]
\everymath{\color{AlgVarColor}}%
\Require prunable weight matrices $\{\mathbf{W}_i\}_{i=1}^{L}$; tile size $(b_1, b_2)$; sparsity target $\rho$; training steps $T$; loss hyperparameters $\lambda_1, \lambda_2$; schedules $\{\tau_t\}, \{\kappa_t\}$
\Ensure learned pruning masks $\{\mathbf{M}_i^{\star}\}$; pruned weights $\{\widehat{\mathbf{W}}_i\}$
\State for every $i$: initialize tile logits $\mathbf{P}_{\text{tile},i}$ with a one-shot prior and 2{:}4 parameters $\mathbf{P}_{\text{2:4},i}$
\For{$t = 1, \dots, T$}
    \For{$i = 1, \dots, L$}
        \State $\tilde{\mathbf{M}}_{\text{tile},i} \gets \mathrm{GS}([\mathbf{P}_{\text{tile},i}, 0]; \tau_t, \kappa_t)_{:,:,0} \otimes \mathbf{1}_{b_1 \times b_2}$ \Comment{soft tile mask}
        \State $\tilde{\mathbf{M}}_{\text{2:4},i} \gets$ Eq.~\ref{eq:2_4_mask} on $\mathbf{P}_{\text{2:4},i}$ \Comment{differentiable 2:4 mask}
        \State $\tilde{\mathbf{M}}_i \gets \tilde{\mathbf{M}}_{\text{tile},i} + (1 - \tilde{\mathbf{M}}_{\text{tile},i}) \odot \tilde{\mathbf{M}}_{\text{2:4},i}$ \Comment{merge masks}
    \EndFor
    \State $\mathcal{L} \gets \mathcal{L}_{LM}(x ; \{\tilde{\mathbf{M}}_i \odot \mathbf{W}_i\}_i) + \lambda_1 \big| \frac{\sum_i \|\tilde{\mathbf{M}}_i\|_1}{\sum_i \|\mathbf{W}_i\|_0} - \rho \big| - \lambda_2 \frac{\sum_i \| \tilde{\mathbf{M}}_i \odot \mathbf{W}_i \|_2^2}{\sum_i \| \mathbf{W}_i \|_2^2}$
    \State update all $\mathbf{P}_{\text{tile},i}$ and $\mathbf{P}_{\text{2:4},i}$ via backpropagation on $\mathcal{L}$ \Comment{Eq.~\ref{eq::loss}}
\EndFor
\For{$i = 1, \dots, L$}
    \State $\mathbf{M}_{\text{tile},i}^{\star} \gets \mathbf{1}[\mathbf{P}_{\text{tile},i} > 0] \otimes \mathbf{1}_{b_1 \times b_2}$; \ $\mathbf{M}_{\text{2:4},i}^{\star} \gets$ best 2:4 mask from $\mathbf{P}_{\text{2:4},i}$ \Comment{binarize}
    \State $\mathbf{M}_i^{\star} \gets \mathbf{M}_{\text{tile},i}^{\star} + (1 - \mathbf{M}_{\text{tile},i}^{\star}) \odot \mathbf{M}_{\text{2:4},i}^{\star}$; \ $\widehat{\mathbf{W}}_i \gets \mathbf{W}_i \odot \mathbf{M}_i^{\star}$
\EndFor
\State \Return $\{\mathbf{M}_i^{\star}\}, \{\widehat{\mathbf{W}}_i\}$
\end{algorithmic}
\end{algorithm}

\paragraph{Memory-efficient \our.} To further reduce overhead, \ours can be run in a memory-efficient manner by freezing the sparse mask parameters and optimizing only the tile-level decisions. \rev{Concretely, a binary 2:4 mask is first obtained, with a one-shot method such as SparseGPT or Wanda or by training MaskLLM, and kept constant, so $\Tilde{\mM}_{2:4}$ in Equation~\ref{eq:tile_merge} becomes a fixed binary mask (Appendix~\ref{app::implementation_details}).} This reduces the number of learnable parameters to \rev{one logit per tile,} $\tfrac{d_1 d_2}{b_1 b_2}$. While this lighter formulation limits mask-selection flexibility and can reduce performance as seen in Table \ref{ablation:tile_only}, it makes training feasible under strict memory constraints, such as fitting an 8B model on a single 80GB GPU. We denote this version of \ours by \ourstile and the joint optimization version of \ours by \oursjoint. 

\section{Efficient deployment of \ours}

Executing \ours requires handling hybrid sparse–dense tiles, a capability not supported by existing GPU libraries. Current tools either focus exclusively on dense computation (e.g., cuBLAS \citep{cublas}, dense CUTLASS \citep{cutlass}, OpenAI Triton \citep{triton}), or restrict support to fixed $2{:}4$ sparsity (e.g., cuSPARSELt \citep{cusparselt}, sparse CUTLASS). STOICC \citep{stoicc} lifts these limitations by extending Triton with hybrid tile-level sparsity, making it a suitable backend for accelerating \ours.

\rev{We distinguish two granularities: \emph{mask tiles}, the $b_1{\times}b_2$ blocks at which \ours learns dense/2:4 decisions (Section~\ref{sec:patch}), and \emph{execution tiles}, the GEMM scheduling blocks that STOICC's autotuner selects and that subdivide mask tiles.} Similar to Triton, STOICC employs an inspector that benchmarks candidate kernel configurations for each sparsity ratio, identifying the most hardware-efficient \rev{execution} tile size for the target GPU. On NVIDIA A100 and A6000 GPUs, our experiments show that the optimal configurations are consistently drawn from $128{\times}128$ or its subdivisions (e.g., $128{\times}64$, $64{\times}128$, $64{\times}64$). In practice, this means that regardless of the sparsity ratio or the layer shape, the chosen $128{\times}128$ \rev{mask-tile} granularity guarantees that STOICC’s autotuned \rev{execution} tiles can be applied consistently. Unless otherwise specified, we adopt \rev{$128{\times}128$ mask tiles} in all \ours experiments. Further implementation details are provided in Appendix~\ref{app::stoicc}.

\section{Experiments}
\label{sec:results}

\paragraph{Model, dataset and evaluation.} We evaluate \ours across diverse transformer architectures, including the Qwen-2.5~\citep{qwen}, Gemma 3~\citep{gemma3}, and LLaMA-2~\citep{llama2} and 3~\citep{llama3} model families, spanning 500M to 13B parameters. Following the dataset size and configurations in MaskLLM \citep{maskllm}, masks are trained for 2000 steps with a batch size of 256 on sequences with a length of 4096 tokens from the SlimPajama dataset~\citep{slimpajama}. 

Following previous LLM compression work \citep{slim, maskllm}, we evaluate the models on eight zero-shot downstream tasks: PIQA~\citep{piqa}, ARC-Easy and ARC-Challenge~\citep{arc}, Winogrande~\citep{winogrande}, OpenBookQA~\citep{openbookqa}, RACE~\citep{race}, HellaSwag~\citep{hellaswag}, and MMLU~\citep{mmlu} using the Language Model Evaluation Harness~\citep{eval-harness} framework. Additionally, similar to previous work \citep{slim, sparsegpt, wanda}, we evaluate the models on a language modeling task using the WikiText2~\citep{wiki} dataset with a sequence length of 4096, comparing against established baselines in the following sections.

\paragraph{Baselines.} To evaluate \ours against established 2:4 sparsity pruning techniques, we compare it with the state-of-the-art learnable method MaskLLM~\citep{maskllm}, as well as one-shot methods including Wanda~\citep{wanda}, SparseGPT~\citep{sparsegpt}, Thanos \citep{thanos}, ProxSparse \citep{proxsparse} and magnitude pruning~\citep{magnitudepruning}. For one-shot pruning methods, following the default configurations in each paper, we prune the models over 128 samples from the C4 dataset.

The publicly available MaskLLM pruned checkpoints are limited to LLaMA-2 7B and LLaMA-3.1 8B models. To ensure a fair comparison across all models, we implemented MaskLLM in PyTorch and replicated its results for additional architectures presented in this study. 

We faced a similar challenge with ProxSparse as well, where only the LLaMA-2-7B and LLaMA-3.1-8B checkpoints are publicly available. We have pruned other models with their official code base using their default hyperparameters for comparison. 

Additional implementation details and hyperparameters used in our experiments are provided in Appendix \ref{app::implementation_details}.

\subsection{Model quality results}

\paragraph{Joint sparse and dense tile optimization.} For smaller models like Qwen-2.5 0.5B, LLaMA-3.2 1B, and Gemma-3 1B, we apply the joint variant \oursjoint, which simultaneously optimizes dense tile locations and sparsity patterns within sparse tiles. This approach enables effective performance.

The average accuracy of the models across eight zero-shot downstream tasks and their perplexity on the WikiText2 dataset is reported in Table \ref{table::joint_patch}. The results demonstrate that \oursjoint provides a flexible tradeoff between sparsity ratio and model quality, narrowing the performance gap to dense models while ensuring hardware-friendly inference. A similar pattern holds for larger models using a memory-efficient variant, as explored next.

\begin{table*}[tb]
\centering
\small
\caption{Model quality (average accuracy across eight zero-shot tasks and perplexity on WikiText2 dataset) for different pruning methods. By jointly optimizing the location of dense tiles and the sparsity pattern within the sparse tiles, \oursjoint allows for a continuous sparsity ratio for the models, providing a flexible tradeoff between sparsity and model quality.}
\setlength{\tabcolsep}{3pt}%
\resizebox{\textwidth}{!}{%
\begin{tabular}{l l l c c c c c c}
\toprule
Sparsity & Method & Pattern &
\multicolumn{2}{c}{Qwen-2.5 0.5B} &
\multicolumn{2}{c}{LLaMA-3.2 1B} &
\multicolumn{2}{c}{Gemma-3 1B} \\
\cmidrule(lr){4-5} \cmidrule(lr){6-7} \cmidrule(lr){8-9}
& & & \taskname{Acc (\% $\uparrow$)} & \taskname{PPL ($\downarrow$)} & \taskname{Acc (\% $\uparrow$)} & \taskname{PPL ($\downarrow$)} & \taskname{Acc (\% $\uparrow$)} & \taskname{PPL ($\downarrow$)} \\
\midrule
0\%  & Dense     & -      & 46.00 & 12.08 & 47.70 & 9.06  & 47.01 & 11.67 \\
\midrule
50\% & Magnitude & 2:4    & 30.16 & 6734.97 & 29.66 & 563.44 & 31.66 & 5005.56 \\
     & Wanda     & 2:4    & 32.97 & 72.48   & 31.61 & 78.18  & 34.16 & 69.41   \\
     & SparseGPT & 2:4    & 34.81 & 36.59   & 35.55 & 32.73  & 35.58 & 44.59   \\
     & Thanos    & 2:4    & 31.31 & 37.32   & 35.71 & 33.03  & 35.09 & 62.63   \\
     & ProxSparse& 2:4    & 32.05 & 111.05   & 33.55 & 49.33  & 36.63 & 90.50 \\
     & MaskLLM   & 2:4    & 39.33 & 15.22   & 41.04 & 12.93  & 41.84 & 12.82   \\
\midrule
\oursrow 45\% & \oursjoint     & Dense/2:4 Tiles & 40.29 & 14.57 & 42.08 & 12.23 & 42.80 & 11.96 \\
\ourssep
\oursrow 35\% & \oursjoint     & Dense/2:4 Tiles & 41.15 & 13.84 & 42.72 & 11.67 & 43.30 & 11.48 \\
\ourssep
\oursrow 25\% & \oursjoint     & Dense/2:4 Tiles & 42.39 & 13.47 & 43.81 & 11.00 & 44.07 & 11.17 \\
\bottomrule
\end{tabular}
}

\label{table::joint_patch}
\end{table*}

\paragraph{Memory-efficient tile selection.} For larger models such as LLaMA-2 7B, LLaMA-2 13B, and LLaMA-3.1 8B, we employ the memory-efficient variant \ourstile, which freezes the fine-grained sparse weight structure while optimizing dense tile selections.

Table \ref{table::tiled_patch} summarizes the average accuracy of the models across eight downstream tasks in addition to their perplexity on the WikiText2 dataset for different sparsity ratios, illustrating that \ourstile delivers a comparable flexible sparsity-quality tradeoff when using a high-quality frozen 2:4 mask. On LLaMA-2 13B, \ourstile reaches eight-task average accuracies of 56.31\%, 54.60\%, and 53.24\% with WikiText2 perplexities of 5.00, 5.44, and 5.85 at 25\%, 35\%, and 45\% sparsity, respectively, demonstrating that \ours scales favorably to larger models. The dense LLaMA-2 13B baseline scores 58.38\% with a perplexity of 4.89, so \rev{\textbf{\ourstile at 25\% sparsity recovers within roughly 2 percentage points of dense accuracy and 0.11 in perplexity}} while still admitting hardware-friendly acceleration. By contrast, the strongest one-shot 2:4 baseline at 50\% sparsity, ProxSparse, reaches only 50.80\% with a perplexity of 7.11, so \ourstile improves over it by 2.4--5.5 accuracy points and 1.3--2.1 in perplexity across the three sparsity ratios, and the remaining one-shot baselines trail by larger margins. MaskLLM is listed as N/A on LLaMA-2 13B, since no public checkpoint exists and its reported training cost exceeds our compute budget (Table~\ref{table::tiled_patch}).

\begin{table*}[tb]
\centering
\small
\caption{Model quality (average accuracy across eight zero-shot tasks and perplexity on WikiText2 dataset) for different pruning methods. By only optimizing the location of dense tiles while keeping sparsity pattern within the sparse tiles frozen, \ourstile provides a memory efficient variant for \oursjoint,  allowing for a continuous sparsity ratio for the models and providing a flexible tradeoff between sparsity and model quality.}
\label{table::tiled_patch}
\setlength{\tabcolsep}{3pt}%
\resizebox{\textwidth}{!}{
\begin{tabular}{l l l c c c c c c}
\toprule
Sparsity & Method & Pattern &
\multicolumn{2}{c}{LLaMA-2 7B} &
\multicolumn{2}{c}{LLaMA-2 13B} &
\multicolumn{2}{c}{LLaMA-3.1 8B} \\
\cmidrule(lr){4-5} \cmidrule(lr){6-7} \cmidrule(lr){8-9}
& & & \taskname{Acc (\% $\uparrow$)} & \taskname{PPL ($\downarrow$)} & \taskname{Acc (\% $\uparrow$)} & \taskname{PPL ($\downarrow$)} & \taskname{Acc (\% $\uparrow$)} & \taskname{PPL ($\downarrow$)} \\
\midrule
0\%  & Dense     & -      & 54.61 & 5.12 & 58.38 & 4.89 & 60.31 & 5.84 \\
\midrule
50\% & Magnitude & 2:4    & 43.44 & 54.39 & 45.94 & 8.89 & 35.93 & 765.92 \\
     & Wanda     & 2:4    & 44.30 & 11.15 & 47.95 & 8.91 & 41.77 & 21.29 \\
     & SparseGPT & 2:4    & 45.09 & 10.12 & 49.67 & 8.86 & 45.53 & 15.11 \\
     & Thanos    & 2:4    & 44.80 & 11.19 & 49.33 & 8.80 & 45.72 & 16.09 \\
     & ProxSparse& 2:4    & 45.92 & 9.18 & 50.80 & 7.11 & 45.14 & 15.17 \\
     & MaskLLM   & 2:4    & 48.62 & 6.78  & N/A$^{\dagger}$ & N/A$^{\dagger}$ & 52.80 & 8.58  \\
\midrule
\oursrow 45\% & \ourstile     & Dense/2:4 Tiles & 48.99 & 6.55 & 53.24 & 5.85 & 53.60 & 8.20 \\
\ourssep
\oursrow 35\% & \ourstile     & Dense/2:4 Tiles & 50.08 & 6.18 & 54.60 & 5.44 & 55.28 & 7.89 \\
\ourssep
\oursrow 25\% & \ourstile     & Dense/2:4 Tiles & 51.58 & 5.86 & 56.31 & 5.00 & 56.48 & 7.34 \\
\bottomrule
\end{tabular}
}
\begin{flushleft}
\footnotesize $^{\dagger}$MaskLLM checkpoints for LLaMA-2 13B are not publicly released, and the authors report a training cost of $\sim$2304 A100 GPU-hours, which exceeds our available compute budget; we therefore omit this baseline on LLaMA-2 13B.
\end{flushleft}

\end{table*}

Overall, across Tables \ref{table::joint_patch} and \ref{table::tiled_patch}, \ours consistently surpasses one-shot methods like Wanda, SparseGPT, and magnitude pruning due to its end-to-end training on large corpora. While MaskLLM also trains end-to-end on a large dataset, its fixed 2:4 sparsity ratio limits achievable accuracy and perplexity. In contrast, \ours overcomes this limitation with flexible dense tile allocation, achieving accuracy gains and perplexity reductions from 45\% to 25\% sparsity that progressively align with dense model performance. The full per-task accuracy results are provided in Appendix \ref{app::per_task}.

\subsection{Understanding the components of \ours}
\label{sec:components}

This subsection examines the design choices driving \our's performance by analyzing its behavior across various configurations on the Qwen-2.5 0.5B model.

\paragraph{Tile size.} We initially assess the impact of \rev{mask} tile size on PATCH's performance, fixing hyperparameters to those optimized for 128$\times$128 tiles. Table~\ref{ablation:tile_size} reveals that $4 \times 4$ tiles maximize model quality through finer sparse-dense control, though larger tile sizes show minimal variation, suggesting robustness. However, smaller tiles may hinder hardware efficiency, requiring a balance with hardware specifications.

\begin{table}[t]
    
    \begin{minipage}{0.50\textwidth}
        \centering%
        \caption{Impact of \our's mask tile size across sparsity levels ($\downarrow$ is better). The effect of tile size on model quality is not significant, showing \our's robustness against tile size.}
\label{ablation:tile_size}
\begin{center}
\begin{small}%
\setlength{\tabcolsep}{4.2pt}%
\begin{tabular}{l c c c c c c}
\toprule
Sparsity & \multicolumn{6}{c}{Mask Tile Size} \\
\cmidrule(lr){2-7}
(0.5B) & 128 & 64 & 32 & 16 & 8 & 4 \\ 
\midrule
45\% & 14.57 & 14.66 & 14.70 & 14.67 & 14.70 & \textbf{14.55} \\
35\% & 13.84 & 14.08 & 14.15 & 14.03 & 14.01 & \textbf{13.72} \\
25\% & 13.47 & 13.54 & 13.52 & 13.53 & 13.40 & \textbf{13.11} \\
\bottomrule
\end{tabular}
\end{small}
\end{center}

    \end{minipage}%
    \hfill%
    \begin{minipage}{0.45\textwidth}
        \centering%
        \caption{Global sparsity yields better quality by concentrating pruning in less important blocks and preserving density elsewhere ($\downarrow$ is better).}
\label{ablation:global_vs_layer_sparsity}
\begin{center}
\begin{small}
\setlength{\tabcolsep}{10pt}%
\begin{tabular}{l c c}
\toprule
Sparsity & Global & Layer-wise \\
(0.5B) & (PPL) & (PPL) \\
\midrule
45\% & \textbf{14.57} & 15.17 \\
35\% & \textbf{13.84} & 14.48 \\
25\% & \textbf{13.47} & 13.95 \\
\bottomrule
\end{tabular}
\end{small}
\end{center}

    \end{minipage}
\end{table}

\paragraph{Joint vs. tile-only mask search.} 
We then analyze the impact of fixing the 2:4 masks and optimizing only tile masks. Table~\ref{ablation:tile_only} shows that among frozen 2:4 masks, MaskLLM provides the strongest results. On the other hand, one-shot pruning methods perform comparably at higher sparsity levels but diverge at lower sparsity, with SparseGPT emerging as the best overall. When comparing against our full approach, joint optimization of both tile and 2:4 masks consistently outperforms tile-only training across sparsity ratios. Nevertheless, tile-only training remains a practical alternative for larger models in resource-constrained settings, as also reflected in Table ~\ref{table::tiled_patch}.

\begin{table}[t]
\caption{Impact of fixed 2:4 mask selection for \ourstile{}, compared with joint optimization ($\downarrow$ is better). \oursjoint{} achieves the lowest perplexity overall, while for \ourstile{}, MaskLLM provides the best frozen mask. \textit{Mag}. refers to Magnitude pruning.}
\label{ablation:tile_only}
\begin{center}
\begin{footnotesize}
\setlength{\tabcolsep}{1.8pt}%
\begin{tabular}{l c c c c | c}
\toprule
\makecell[l]{Sparsity \\ (0.5B)} & MaskLLM & \makecell[c]{SparseGPT \\ (no update)} & Wanda & Mag. & \oursjoint \\
\midrule
45\% & \textbf{15.06} & 21.84 & 21.83 & 21.33 & 14.57 \\
35\% & \textbf{14.55} & 17.29 & 17.96 & 19.90 & 13.84 \\
25\% & \textbf{14.17} & 14.89 & 15.09 & 16.05 & 13.47 \\
\bottomrule
\end{tabular}
\end{footnotesize}
\end{center}
\vskip -0.1in
\end{table}

\begin{figure}[tb]
\centering
\includegraphics[width=0.70\linewidth]{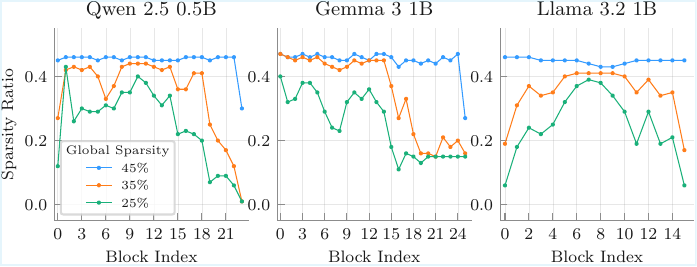}
\caption{Layer-wise sparsity allocation under different global sparsity budgets for various models. \ours achieves the target global sparsity while flexibly distributing pruning across transformer layers.}
\label{fig:sparsity_allocation}
\end{figure}

\paragraph{Sparsity allocation.} We analyze how sparsity is allocated across transformer blocks under a global target. Across models, deeper transformer blocks are pruned far less, while the initial blocks also tend to receive lighter pruning depending on the architecture. By contrast, the middle blocks consistently absorb most of the sparsity, suggesting that they contain more redundancy (Figure~\ref{fig:sparsity_allocation}). We compare this flexible allocation to enforcing sparsity uniformly at the layer level. As shown in Table~\ref{ablation:global_vs_layer_sparsity}, global targets deliver better results by pruning more aggressively in redundant layers while preserving capacity in sensitive ones. In contrast, layer-wise targets impose uniform sparsity that can over-prune critical components \citep{adaptive1, adaptive2, adaptive3, owl}. \rev{\textbf{The learned allocation, not merely the lower sparsity, drives the gains}: heuristic allocations inspired by \our's learned patterns, which keep attention and the final blocks dense and apply 2:4 to the FFNs with SparseGPT or even MaskLLM masks, reach lower accuracy than \ours at comparable or lower sparsity on every evaluated model, at essentially identical throughput (Appendix~\ref{app::ffn_only}).}

On top of variation across depth, sparsity is also distributed unevenly across the individual linear layers within each transformer block. Figure~\ref{fig:linear_layer_sparsity_allocation_qwen} \rev{(Appendix~\ref{app::sparsity_distribution})} breaks down the allocation into the query, key, value, and output matrices of the attention module, as well as the up, gate, and down matrices of the MLP for the Qwen 2.5 0.5B model. The up, gate, and down layers absorb most of the sparsity and largely explain the overall allocation pattern seen in Figure~\ref{fig:sparsity_allocation}. In contrast, the attention module is treated as more critical. The key and value matrices are never pruned, while the output matrix shows moderate pruning at higher global sparsity targets. The query matrix is pruned the most, suggesting it is the least important within the attention submodule. The distributions for the Gemma-3-1B and LLaMA-3.2-1B models are provided in Appendix \ref{app::sparsity_distribution}, where the same pattern is observed.

\subsection{Combination with other compression methods}
\label{sec:compression_methods}

LLM compression relies on three orthogonal methods, sparsity, quantization, and low-rank approximation, which can be combined. While this work focuses on sparsity, this section demonstrates how \ours integrates with these other techniques. 

\paragraph{Quantization.} Quantization reduces memory and accelerates computation by lowering numerical precision on hardware optimized for low bitwidths.

\paragraph{Low-rank approximation.} Low-rank methods complement sparsity and quantization by reintroducing a small number of parameters to recover accuracy, with \textsc{SLiM} \citep{slim} as a leading one-shot technique.

Table~\ref{table::quantization} reports results on LLaMA-2 7B and LLaMA-3.1 8B, comparing \ours and MaskLLM under 4-bit weight-only quantization, plus a setting that adds a low-rank adapter (10\% of the weight's rank). The results show that sparsity, quantization, and low-rank approximation compose into controllable tradeoffs between compression and quality, and that \ours integrates seamlessly with both within broader compression pipelines.

\begin{table*}[tb]
\centering
\caption{Average accuracy ($\uparrow$ indicates better) across eight zero-shot downstream tasks and WikiText2 perplexity ($\downarrow$ indicates better) of compressed models with \textbf{4-bit weight-only quantization}. Please note that using LoRA adds additional parameters to the model. \textit{Comp. Ratio} refers to the theoretical weight memory compression factor relative to the dense model\rev{; LoRA adapters are quantized to 4-bit integers}.}
\label{table::quantization}
\footnotesize
\setlength{\tabcolsep}{4pt}
\begin{tabular}{l l l l l c c c c}
\toprule
Sparsity & Method & Bit & LoRA &
\multicolumn{2}{c}{LLaMA-2-7B} &
\multicolumn{2}{c}{LLaMA-3.1-8B} & \taskname{Comp.} \\
\cmidrule(lr){5-6} \cmidrule(lr){7-8}
&  & & & \taskname{Acc (\% $\uparrow$)} & \taskname{PPL ($\downarrow$)} & \taskname{Acc (\% $\uparrow$)} & \taskname{PPL ($\downarrow$)} & \taskname{Ratio} \\
\midrule
0\%  & Dense     & -      & -      & 54.61 & 5.12 & 60.31 & 5.84 & 1x \\
\midrule
50\% & MaskLLM   & 4    & -      & 47.98 & 7.64 & 51.12 & 9.92 & 5.33x \\
\midrule
\oursrow
45\% & \ourstile     & 4 & -      & 48.19 & 7.34 & 52.47 & 9.68 & 5.16x \\
\ourssep
\oursrow
45\% & \ourstile     & 4 & \textsc{SLiM}-LoRA & 50.71 & 6.83 & 54.04 & 9.12 & 4.10x \\
\ourssep
\oursrow
35\% & \ourstile     & 4 & -      & 49.38 & 6.92 & 53.81 & 9.26 & 4.85x\\
\ourssep
\oursrow
35\% & \ourstile     & 4 & \textsc{SLiM}-LoRA & 51.91 & 6.42 & 55.70 & 8.37 & 3.90x \\
\ourssep
\oursrow
25\% & \ourstile     & 4 & -      & 50.45 & 6.57 & 55.45 & 8.69 & 4.57x \\
\ourssep
\oursrow
25\% & \ourstile     & 4 & \textsc{SLiM}-LoRA & 52.62 & 6.11 & 56.99 & 7.77 & 3.72x \\
\bottomrule
\end{tabular}
\end{table*}

\subsection{Speedup and memory savings}
\label{sec:speedup}

\begin{figure}[t]
\centering
\includegraphics[width=\linewidth]{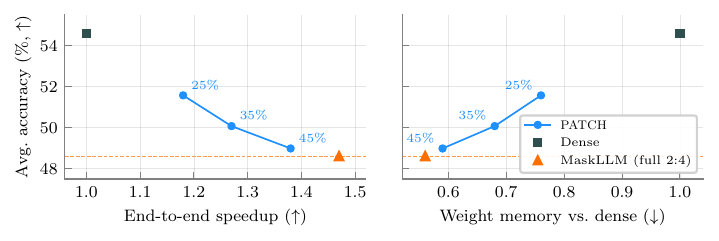}
\caption{\rev{Accuracy--efficiency trade-off on LLaMA-2 7B (A6000, batch size 16, prefill and generation length 128). Eight-task average accuracy versus end-to-end speedup over dense (left) and versus weight memory relative to dense (right). Dense and full 2:4 (MaskLLM, speedup measured with STOICC) are single operating points; \ours at 25\%, 35\%, and 45\% sparsity fills the region between them, and every \ours point lies above the full-2:4 accuracy (dashed).}}
\label{fig:tradeoff}
\end{figure}

We evaluate the inference efficiency of LLaMA-2 7B pruned with \ours using the STOICC \citep{stoicc} compiler. With a batch size of 16 on an A6000 GPU, end-to-end throughput improves by 1.18$\times$, 1.27$\times$, and 1.38$\times$ at 25\%, 35\%, and 45\% sparsity, while \rev{weight} memory drops to 0.76$\times$, 0.68$\times$, and 0.59$\times$ of the dense model\rev{. Figure~\ref{fig:tradeoff} places these operating points next to dense and full 2:4 (MaskLLM), whose speedup of 1.47$\times$ (1.40$\times$ with cuSPARSELt) and 0.56$\times$ weight memory are set by the sparsity format rather than by how the mask is selected. \textbf{\ours fills the previously empty, hardware-realizable interior of the accuracy--speedup curve between dense and full 2:4}: relative to the full-2:4 ceiling, \ours at 45\% gives up 0.09$\times$ of speedup for +0.37 accuracy points, and at 25\% gives up 0.29$\times$ for +2.96. These gains come from extra density by design, and \our's contribution is learning where to allocate it: heuristic allocations of the same density reach lower accuracy at essentially identical throughput (Section~\ref{sec:components}, Appendix~\ref{app::ffn_only}). Because \ours prunes only weights, the KV cache and activations are unchanged, and measured peak memory falls from 16.7\,GB (dense) to 11.8\,GB at 45\% (Appendix~\ref{app::memory}). Appendix~\ref{app::stoicc} extends these measurements to the A100 and LLaMA-2 13B, and Appendix~\ref{app::tile_speedup} to long contexts with FlashAttention-2~\citep{flashattention2}.}

\section{Related work}
\label{sec:related_work_main}

\begin{revblock}
We give a very short discussion of the most relevant work here; a more comprehensive discussion is provided in Appendix~\ref{sec:related_work}. Within the sparsity pillar of the Compression Trinity~\citep{trinity_hub, trinity_thesis}, SparseGPT~\citep{sparsegpt}, Thanos~\citep{thanos}, OPTIMA~\citep{optima}, and Wanda~\citep{wanda} prune LLMs layer by layer in one shot, while MaskLLM~\citep{maskllm} and ProxSparse~\citep{proxsparse} learn 2:4 masks end-to-end, and LEAP~\citep{leap} extends end-to-end mask learning to unstructured sparsity. The 2:4 methods are confined to a uniform 50\% pattern; \ours extends MaskLLM's formulation with learnable dense tiles. Allocation methods such as OWL~\citep{owl} and AlphaPruning~\citep{alphapruning} vary sparsity across layers but target unstructured sparsity, which GPUs accelerate poorly~\citep{flashllm, spinfer}; \ours allocates sparsity at tile granularity within the accelerated 2:4 format and composes with the quantization and low-rank pillars (Section~\ref{sec:compression_methods}).
\end{revblock}

\section{Limitations}
\label{sec:limitations}

\begin{revblock}
The main limitation of \ours is the cost of learning masks end-to-end. Joint optimization of tile allocations and fine-grained 2:4 patterns requires iterative sampling and backpropagation through the full model, taking 18 to 160 H100 GPU-hours for the 0.5B to 13B models (Appendix~\ref{app::implementation_details}), far more than one-shot pruning. This is a one-time cost amortized over every subsequent inference, and \ourstile reduces it by freezing the 2:4 structure, but improving training efficiency remains important future work. Second, because each tile is either dense or 2:4, the global sparsity of \ours is bounded by 50\%; higher ratios require additional patterns such as 1:4 (Appendix~\ref{app::pattern_agnostic}). Third, the realized speedup depends on the kernel backend: STOICC's gains are narrower on A100 than on A6000 (1.16$\times$ versus 1.38$\times$ at 45\% sparsity on LLaMA-2 7B), kernels for newer architectures (Hopper, Blackwell) and low-precision formats such as NVFP4 are not yet available, and at long prefills attention dominates the runtime, reducing the 45\% speedup to 1.22$\times$ at 8K and 1.14$\times$ at 16K tokens (Appendix~\ref{app::tile_speedup}). Finally, masks are learned from a task-agnostic language-modeling objective, so a single allocation serves all downstream tasks, and knowledge-intensive tasks such as MMLU degrade the most (Appendix~\ref{app::per_task}).
\end{revblock}

\section{Conclusion}

We introduced \ours, a hybrid sparsity framework that partitions weight matrices into tiles designated as dense or 2:4 sparse, enabling adaptive sparsity between 0\% and 50\%. Across models up to 13B parameters, \ours consistently improves accuracy over state-of-the-art 2:4 pruning while delivering up to 1.38$\times$ end-to-end speedup on consumer-grade GPUs, motivating future work on broader sparsity formats, quantization, and hardware co-design.

\clearpage

\section*{Societal impact}
\label{sec:societal_impact}

This work lowers the computational and energy costs of large language models, helping make efficient model deployment more accessible across resource-constrained settings and devices. By reducing reliance on large-scale infrastructure, it has the potential to democratize access to advanced models and improve the sustainability of AI systems. At the same time, broader accessibility raises risks related to misuse and performance degradation in high-stakes applications. These considerations highlight the need for careful evaluation, transparency, and safeguards to ensure efficiency gains are paired with responsible deployment.

\section*{AI use statement}
\label{sec:ai_statement}

We used language models to enhance the readability of the manuscript, correct grammatical and typographical errors, and ensure conformity with the NeurIPS author guidelines. Beyond their application in benchmark evaluations and experimental procedures, language models were not employed in any other aspect of this study.

\section*{Reproducibility statement}
\label{sec:reproducibility}

\begin{revblock}
Our implementation, training scripts, and evaluation pipeline are available at \url{https://github.com/Paramathic/patch}, and the pruned checkpoints are released at \url{https://huggingface.co/collections/mohammad-mozaffari/patch} as binary masks that apply to the public base models, each under its base model's license. \ours is specified in Section~\ref{sec:patch} and Algorithm~\ref{alg:patch}, and the experimental setup in Section~\ref{sec:results}. All models are publicly available checkpoints, and all mask-training data (SlimPajama), one-shot calibration data (128 C4 samples), and evaluation data are public. For MaskLLM and ProxSparse, public checkpoints exist only for LLaMA-2 7B and LLaMA-3.1 8B; for the remaining models we reproduce MaskLLM with our PyTorch implementation and ProxSparse with its official code base and default hyperparameters (Section~\ref{sec:results}). Every \ours hyperparameter, the training budget, and the GPU-hour cost per model are listed in Appendix~\ref{app::implementation_details}, and robustness across seeds and calibration corpora is reported in Appendices~\ref{app::seed} and~\ref{app::c4_vs_slimpajama}. Masks are trained on a single node with 4 NVIDIA H100 GPUs; inference throughput is measured on NVIDIA A6000 and A100 GPUs with the STOICC compiler~\citep{stoicc} (Appendix~\ref{app::stoicc}). Software stack: \texttt{lm-evaluation-harness v0.4.9.1}~\citep{eval-harness}, \texttt{torch v2.7.1+cu126}, \texttt{transformers v4.57.0}, and \texttt{accelerate v1.10.1}.
\end{revblock}

\begin{ack}
This work was supported in part by NSERC Discovery Grants (RGPIN-06516, DGECR00303), the Canada Research Chairs program, the Ontario Early Researcher Award, and the Digital Research Alliance of Canada (\url{www.alliancecan.ca}). We extend our gratitude towards Ray Hung for assistance with the results, and Victor Kamel and Arya Rafii for their help in integrating STOICC into our work.
\end{ack}

\clearpage
\bibliographystyle{plainnat}
\bibliography{neurips_2026}

\clearpage
\appendix
{\LARGE\bfseries Appendix\par}
\vspace{0.8em}
{\large\bfseries Table of Contents\par}
\vspace{0.3em}
\hrule height 0.8pt
\vspace{0.5em}
\startcontents[appendix]
\printcontents[appendix]{}{1}{\setcounter{tocdepth}{2}}
\vspace{0.5em}
\hrule height 0.8pt
\vspace{1em}

\section{Related work}
\label{sec:related_work}

\begin{revblock}
Model compression research divides across three pillars: sparsity, quantization, and low-rank approximation, a triad referred to as the \emph{Compression Trinity}~\citep{trinity_hub, trinity_thesis}. The pillars are not siloed: techniques proven in one repeatedly transfer to another, and they compose within a single pipeline, as Section~\ref{sec:compression_methods} shows for \ours with 4-bit quantization and low-rank adapters. We first survey the sparsity pillar in depth (Section~\ref{app:rw_pruning}), then the complementary pillars (Section~\ref{app:rw_pillars}).
\end{revblock}

\subsection{Pruning methods}
\label{app:rw_pruning}

Pruning is one of the most widely studied approaches for compressing deep neural networks, with the goal of removing redundant parameters while preserving accuracy. Classical pruning methods can be broadly categorized into \emph{local} (layer-wise) and \emph{global} (end-to-end) strategies.

\paragraph{Local pruning.}
Local approaches prune each layer independently, typically by minimizing reconstruction error within that layer. A seminal example is Optimal Brain Surgeon (OBS) \citep{obs}\rev{, extended to layer-wise compression by Optimal Brain Compression (OBC) \citep{obc}}, which leverages second-order information to identify and remove weights while updating the remaining parameters to compensate for loss. While highly principled, the quadratic cost of computing and inverting the Hessian makes OBS infeasible for large models. 

Recent work adapts these ideas to LLM-scale pruning. SparseGPT \citep{sparsegpt} formulates layer-wise pruning as a sparse regression problem, enabling efficient approximations of OBS that scale to billion-parameter models. Thanos \citep{thanos} further improves accuracy by employing multi-column approximations to reduce error accumulation. Wanda \citep{wanda}, on the other hand, discards explicit weight updates and instead uses a simple magnitude-activation criterion with calibration data, yielding competitive quality with extremely fast runtimes. \rev{OPTIMA \citep{optima} casts the weight reconstruction that follows mask selection as independent row-wise quadratic programs sharing a common layer Hessian, yielding per-row optimal updates.} Despite their efficiency, local methods often suffer from limited capacity to recover accuracy since pruning decisions ignore cross-layer dependencies.

\paragraph{Global pruning.}
Global approaches aim to jointly optimize pruning decisions across layers, typically leading to better overall trade-offs. Optimal Brain Damage (OBD) \citep{obd} is an early global method that estimates weight saliency using the diagonal Hessian. Extensions such as WoodFisher \citep{woodfisher} approximate the Hessian via Kronecker factorizations, making computation more tractable but still challenging for modern LLMs\rev{; MKOR \citep{mkor} further reduces the cost of Kronecker-factored second-order approximations by maintaining them with rank-1 updates, though for training rather than pruning}. 

More recent approaches bypass costly second-order computations. MaskLLM \citep{maskllm} formulates pruning as a binary classification task (keep vs.\ prune) and solves it using standard optimizers such as AdamW \citep{adamw}, achieving strong results even under hardware-friendly structured sparsity (e.g., $2{:}4$). ProxSparse \citep{proxsparse} instead adopts a proximal regularization framework, reducing the overhead of MaskLLM while trading off some pruning accuracy. \rev{LEAP \citep{leap} extends learnable masking beyond fixed patterns, making end-to-end unstructured mask learning tractable through a per-weight Bernoulli relaxation via Gumbel-sigmoid.} These works highlight the tension between pruning quality and efficiency: global methods often achieve higher accuracy but remain more computationally expensive than simple one-shot local pruning.

\subsection{Complementary compression pillars}
\label{app:rw_pillars}

Beyond pruning, several orthogonal compression techniques are widely used and can be combined with sparsity for additional gains. \emph{Quantization} reduces the bit precision of parameters and activations, e.g., from 32-bit floating point to 8- or 4-bit integers, thereby reducing memory footprint and accelerating inference \citep{quantization_survey, quantization_survey2}. 

\emph{Low-rank adaptation} methods decompose weight matrices into smaller factors, effectively reducing parameter counts while maintaining expressivity. Recent approaches such as LQ-LoRA \citep{lqlora}, SLiM \citep{slim}, and SLoPe \citep{slope} demonstrate that low-rank structures can be used both for efficient fine-tuning and for direct model compression. \rev{The pillars also compose: SLiM performs one-shot joint sparse-plus-quantized compression with a mathematically computed low-rank correction, demonstrating cross-pillar gains in a single pass \citep{slim}, and SLoPe combines sparse pretraining with lazy low-rank adapters introduced in the final iterations of pretraining \citep{slope}.} 

Finally, \emph{knowledge distillation} \citep{knowledge_distillation} transfers knowledge from a large teacher model to a smaller student, yielding compact models that retain much of the teacher’s performance. These methods are complementary to pruning, and hybrid frameworks that integrate sparsity, quantization, and low-rank factorization represent a promising direction for achieving high compression ratios without sacrificing accuracy.

\section{STOICC integration}
\label{app::stoicc}

Triton \citep{triton} enables developers to write efficient GPU kernels with a Python-like syntax, but it natively supports only dense matrix operations and cannot handle sparsity. To accelerate the mixed-tile format produced by \our, we employ the STOICC compiler \citep{stoicc}. STOICC extends Triton with a sparse code-generation backend that allows tiles within a matrix to be either dense or sparse, enabling mixed execution within a single matrix multiplication.

We rely on STOICC’s inspector to autotune both \rev{execution} tile sizes and execution schedules (i.e., alternative kernel execution schemes such as split-$K$ parallelism) for the prefill and decoding stages of LLM inference. Matrix compression and metadata generation are determined by the chosen \rev{execution} tile size, which must remain consistent across both stages. To address this, we first autotune the decoding stage, which is the primary bottleneck of autoregressive generation, since it is executed once per generated token (e.g., 128 times for 128 new tokens), unlike the single pass of prefill. The optimal \rev{execution} tile size identified for decoding is then fixed and reused for prefill, where we perform a second round of autotuning over the remaining independent parameters. 

In contrast, for fully 2:4 sparse matrices, compression is independent of the block size, so they can be autotuned in the same way as dense kernels in Triton without this coupling constraint.

The pseudocode outlining this process, including the handling of dense, fully 2:4 sparse, and mixed-sparsity modules, is provided in PseudoCode~\ref{code:stoicc_pseudo}.

\definecolor{keywordcolor}{rgb}{0.6, 0.1, 0.6}%
\definecolor{commentcolor}{rgb}{0.4, 0.4, 0.4}
\definecolor{stringcolor}{rgb}{0.1, 0.5, 0.1}
\definecolor{backgroundcolor}{rgb}{0.95, 0.95, 0.95}

\definecolor{classcolor}{rgb}{0.45, 0.15, 0.65}%
\definecolor{funccolor}{rgb}{0.10, 0.35, 0.80}%
\definecolor{varcolor}{rgb}{0.00, 0.00, 0.00} 
\definecolor{rulecolor}{gray}{0.25}

\lstdefinelanguage{Pseudo}{
  morekeywords={for,each,in,if,elif,else,continue,return,where},
  sensitive=true,
  morecomment=[l]{//},
  morestring=[b]"
}

\lstdefinestyle{PseudoStyle}{
    backgroundcolor=\color{backgroundcolor},
    basicstyle=\ttfamily\small,
    language=Pseudo,
    keywordstyle=\color{keywordcolor}\bfseries,
    commentstyle=\color{commentcolor}\itshape,
    stringstyle=\color{stringcolor},
    showstringspaces=false,
    frame=single,
    rulecolor=\color{rulecolor},
    numbers=none,
    numberstyle=\tiny\color{gray},
    numbersep=6pt,
    breaklines=true,
    tabsize=2,
    keepspaces=true,
    captionpos=b,
    emph={Inspector,MixedModule,Tensor, STOICC},%
    emphstyle=\color{classcolor}\bfseries,
    emph={[2]create_configs,select_2_4_backend,inspect,compress,
          create_module,replace,set_configs, get_sparsity_ratio},
    emphstyle={[2]\color{funccolor}},
    moredelim=**[is][\color{varcolor}]{|}{|},%
     mathescape=true,
    escapeinside={(*@}{@*)},
    alsoletter={\in},
    xleftmargin=5pt,
    xrightmargin=5pt,
}

\begingroup
\renewcommand{\lstlistingname}{PseudoCode}

\begin{lstlisting}[style=PseudoStyle,
  caption={Tuning and Converting Model Weights to Mixed Format.},
  label={code:stoicc_pseudo}]
def tune_and_convert_model(|M|, |backend_name|):   
    // backend_name (*@ {\color{commentcolor}$\in$} @*) {"STOICC", "cuSPARSELt"}
    |2_4_backend| = select_2_4_backend(|backend_name|)

    // create all configs & schedules to tune over
    |base_configs| = STOICC.create_configs()
    |inspector|    = Inspector()

    for each |module| in |M|:
        |s| = get_sparsity_ratio(|module.weight|)

        // Keep dense Torch (cuBLAS) module
        if |s| == 0:
            continue

        // Use STOICC or cuSPARSELt for fully 2:4
        elif |s| == 0.5:
            |c|          = |2_4_backend|.compress(|module.weight|)
            |new_module| = |2_4_backend|.create_module(|c|)
            replace(|module|, |new_module|)
            continue

        else:
            |decoding_input| = Tensor(|BS|, |module.weight.shape[1]|)
            |prefill_input|  = Tensor(|BS| * |SL|, |module.weight.shape[1]|)

            // Tune on decoding input first
            |inspector|.set_configs(|base_configs|)
            |best_cfg_dec| = |inspector|.inspect(
                |decoding_input|, 
                |module.weight|, 
                isASparse=False)
            |BN| = |best_cfg_dec|["BLOCK_N"]
            |BK| = |best_cfg_dec|["BLOCK_K"]

            // Tune on prefill using decoding tile sizes
            |prefill_cfg| = STOICC.create_configs(BLOCK_N=|BN|, BLOCK_K=|BK|)
            |inspector|.set_configs(|prefill_cfg|)
            |best_cfg_pre| = |inspector|.inspect(
                |prefill_input|, 
                |module.weight|, 
                isASparse=False)

            |c| = |inspector|.compress(|module.weight|, |BN|, |BK|)
            |mixed_module| = MixedModule(|c|, |best_cfg_dec|, |best_cfg_pre|)
            replace(|module|, |mixed_module|)

    return |M|
\end{lstlisting}

\endgroup

Table~\ref{table::speedup} reports the measured throughput (tokens processed per second) of LLaMA-2 7B at sparsity levels of 45\%, 35\%, and 25\% with a batch size of 16 on an A6000 GPU. To reduce CPU overhead from launching Triton kernels in PyTorch, we executed generation through CUDA graphs, capturing both the prefill and decoding stages. With sparsity ratios between 25\% and 45\%, our heterogeneous approach achieves 1.18$\times$–1.38$\times$ end-to-end acceleration over the dense baseline. We also report timings on A100 in Table~\ref{table::speedup_a100}.

\begin{revblock}
\paragraph{Narrower gains on A100.} The smaller end-to-end speedups on A100 stem from two factors, both external to \our's sparsity format. First, workload scale: at the benchmarked setting (LLaMA-2 7B, batch size 16), the GEMM shapes are too small to fully exploit the A100's higher memory bandwidth and Tensor Core throughput. The resulting kernels expose limited parallelism relative to the A100's resources, shifting execution toward a launch- and latency-dominated regime in which the byte and FLOP savings from sparsity translate into smaller wall-clock gains; on the A6000, whose lower bandwidth places the same shapes closer to the bandwidth-bound regime, the reduction in bytes read pays off more directly. Consistent with this explanation, the larger LLaMA-2 13B reaches 1.27$\times$ and 1.21$\times$ speedups on A100 at 45\% and 35\% sparsity under the same setting (Table~\ref{table::speedup_a100}), about 0.10$\times$ more than the 7B model. Second, backend maturity: STOICC's generated kernels currently achieve smaller per-layer (GEMM-level) speedups on A100 than on A6000, and this gap carries through to the end-to-end numbers. Improving STOICC's code generation for datacenter GPUs is outside the scope of this work; our goal is to show that hybrid tile-level sparsity can be accelerated in practice on both consumer-grade (A6000) and datacenter-grade (A100) hardware.
\end{revblock}

\begin{revblock}
\paragraph{Accuracy, speedup, and memory together.} Table~\ref{table::tradeoff} combines the accuracy, speedup, and weight-memory results for LLaMA-2 7B (plotted in Figure~\ref{fig:tradeoff}). Throughput of a full 2:4 model is determined by the sparsity format, not by how the mask was selected, so the full-2:4 throughput measured in Table~\ref{table::ffn_only_speedup} applies to MaskLLM as well. Relative to this full-2:4 ceiling, \ours at 45\% gives up 0.09$\times$ of speedup for +0.37 accuracy points, and at 25\% gives up 0.29$\times$ for +2.96.
\end{revblock}

\begin{table}[h]
\revcolor
\centering
\caption{Accuracy, speedup, and weight memory of LLaMA-2 7B on an A6000 GPU (batch size 16, prefill length 128, generation length 128). Throughput of a full 2:4 model is determined by the sparsity format, so the full-2:4 throughput of Table~\ref{table::ffn_only_speedup} applies to MaskLLM; its speedup is given as STOICC\,/\,cuSPARSELt.}
\label{table::tradeoff}
\small
\begin{tabular}{l c c c c}
\toprule
Method & Sparsity & \taskname{Avg Acc (\% $\uparrow$)} & \taskname{Speedup ($\uparrow$)} & \taskname{Weight memory ($\downarrow$)} \\
\midrule
Dense & 0\% & 54.61 & 1.00$\times$ & 1.00$\times$ \\
MaskLLM (full 2:4) & 50\% & 48.62 & 1.47$\times$\,/\,1.40$\times$ & 0.56$\times$ \\
\midrule
\oursrow \ours & 45\% & 48.99 & 1.38$\times$ & 0.59$\times$ \\
\ourssep
\oursrow \ours & 35\% & 50.08 & 1.27$\times$ & 0.68$\times$ \\
\ourssep
\oursrow \ours & 25\% & 51.58 & 1.18$\times$ & 0.76$\times$ \\
\bottomrule
\end{tabular}
\end{table}

\begin{table}[h]
\centering
\caption{Throughput of LLaMA-2 7B with mixed sparsity compared to the dense model. Measurements taken on an A6000 GPU with batch size 16. Throughput is reported in tokens processed/sec.}
\label{table::speedup}
\begin{tabular}{ccccc}
\toprule
Sparsity & Prefill length & Tokens generated & Throughput (tok/s) & Speedup vs. dense \\
\midrule
0\%  & 128 & 128  & 1023.80 & 1.00$\times$ \\
25\% & 128 & 128  & 1212.79 & 1.18$\times$ \\
35\% & 128 & 128  & 1304.46 & 1.27$\times$ \\
45\% & 128 & 128  & 1410.20 & 1.38$\times$ \\
\midrule
0\%  & 128 & 1024 & 435.42  & 1.00$\times$ \\
25\% & 128 & 1024 & 493.33 & 1.13$\times$ \\
35\% & 128 & 1024 & 515.39 & 1.18$\times$ \\
45\% & 128 & 1024 & 542.87 & 1.25$\times$ \\
\bottomrule
\end{tabular}
\end{table}

\begin{table}[tb]
\centering
\caption{Throughput of LLaMA-2 7B \rev{and LLaMA-2 13B} with mixed sparsity compared to the dense model. Measurements taken on an A100 GPU with batch size 16. Throughput is reported in tokens processed/sec.}
\label{table::speedup_a100}
\begin{tabular}{cccccc}
\toprule
Model & Sparsity & Prefill & Generated & Throughput (tok/s) & Speedup vs.\ dense \\
\midrule
\multirow{8}{*}{LLaMA-2 7B}
& 0\%  & 128 & 128  & 1876.24& 1.00$\times$ \\
& 25\% & 128 & 128  & 2002.02& 1.07$\times$\\
& 35\% & 128 & 128  & 2088.98& 1.11$\times$\\
& 45\% & 128 & 128  & 2180.88& 1.16$\times$\\
\cmidrule{2-6}
& 0\%  & 128 & 1024 & 812.55& 1.00$\times$ \\
& 25\% & 128 & 1024 & 864.66& 1.06$\times$\\
& 35\% & 128 & 1024 & 885.90& 1.09$\times$\\
& 45\% & 128 & 1024 & 907.12& 1.12$\times$\\
\midrule
\multirow{3}{*}{\rev{LLaMA-2 13B}}
& \rev{0\%}  & \rev{128} & \rev{128} & \rev{1198.98} & \rev{1.00$\times$} \\
& \rev{35\%} & \rev{128} & \rev{128} & \rev{1445.48} & \rev{1.21$\times$} \\
& \rev{45\%} & \rev{128} & \rev{128} & \rev{1519} & \rev{1.27$\times$} \\
\bottomrule
\end{tabular}
\end{table}

\section{Memory analysis}
\label{app::memory}

\begin{revblock}
\ours prunes weights only, so the KV cache and activation memory at inference are identical to those of the dense model, and the entire reduction comes from weight storage. Table~\ref{table::memory} breaks down peak GPU memory for the benchmarked setting of Section~\ref{sec:speedup} (LLaMA-2 7B, FP16, batch size 16, prefill length 128).

\paragraph{Weights.} The dense model occupies $\approx$13.5\,GB in FP16. \ours reduces model memory to 0.76$\times$, 0.68$\times$, and 0.59$\times$ of dense at 25\%, 35\%, and 45\% sparsity, i.e., $\approx$10.3, 9.2, and 8.0\,GB.

\paragraph{KV cache.} Each token stores $2$ (K and V) $\times\, 32$ layers $\times\, 4096$ hidden dimensions $\times\, 2$ bytes $\approx 0.5$\,MB. At batch size 16, this gives $\approx$2\,GB for 128 prefill + 128 generated tokens and $\approx$9\,GB for 128 + 1024, identical across all sparsity levels. Since the cache is pre-allocated to the full sequence length at CUDA-graph capture, peak usage coincides with the prefill phase.

\paragraph{Activations.} Inference retains no autograd state, so activations reduce to transient buffers of the layer currently executing, sized by the tokens in flight (batch $\times$ prefill $= 2048$ during prefill, 16 during decoding). The largest such buffers are the MLP intermediates ($2048 \times 11008 \times 2$\,B $\approx$ 45\,MB each) and the output logits ($\approx$130\,MB when materialized over the full prefill), giving a peak below $\approx$0.5\,GB, one to two orders of magnitude smaller than the weights and KV cache, and likewise unchanged by sparsity. The measured peaks in Table~\ref{table::memory} confirm this analysis.

\paragraph{Training memory.} Our training protocol increases the per-GPU batch size until device memory is saturated, so peak training memory is by construction close to capacity and is not informative on its own. The informative facts are that mask training for all models except LLaMA-2 13B fits on a single 80\,GB GPU, and that the 13B model trains on a single 4-GPU node with a local batch size larger than 1, indicating available headroom.
\end{revblock}

\begin{table}[h]
\revcolor
\centering
\caption{Peak GPU memory breakdown of LLaMA-2 7B in FP16 at inference (A6000, batch size 16, prefill length 128). \ours reduces only the weights; the KV cache and transient activations are identical to the dense model. Analytical peaks sum the components; empirical peaks are measured.}
\label{table::memory}
\footnotesize
\setlength{\tabcolsep}{3.5pt}
\begin{tabular}{l c c c c}
\toprule
Component & Dense & \ours 25\% & \ours 35\% & \ours 45\% \\
\midrule
Weights & 13.5\,GB (1.00$\times$) & $\approx$10.3\,GB (0.76$\times$) & $\approx$9.2\,GB (0.68$\times$) & $\approx$8.0\,GB (0.59$\times$) \\
KV cache (128 generated) & $\approx$2\,GB & $\approx$2\,GB & $\approx$2\,GB & $\approx$2\,GB \\
KV cache (1024 generated) & $\approx$9\,GB & $\approx$9\,GB & $\approx$9\,GB & $\approx$9\,GB \\
Transient activations & $<$0.5\,GB & $<$0.5\,GB & $<$0.5\,GB & $<$0.5\,GB \\
\midrule
Analytical peak (128 generated) & $\approx$16.0\,GB & $\approx$12.8\,GB & $\approx$11.7\,GB & $\approx$10.5\,GB \\
Empirical peak (128 generated) & 16.721\,GB & 14.189\,GB & 12.530\,GB & 11.769\,GB \\
\midrule
Analytical peak (1024 generated) & $\approx$23.0\,GB & $\approx$19.8\,GB & $\approx$18.7\,GB & $\approx$17.5\,GB \\
Empirical peak (1024 generated) & 23.457\,GB & 20.946\,GB & 19.684\,GB & 18.556\,GB \\
\bottomrule
\end{tabular}
\end{table}

\section{Per-task results}
\label{app::per_task}

This appendix provides detailed per-task accuracy results for the models evaluated in Section \ref{sec:results}, covering eight zero-shot downstream tasks: MMLU, PIQA, ARC-Easy, ARC-Challenge, Winogrande, OpenbookQA, RACE, and Hellaswag. The results are presented for each model at various sparsity levels and pruning methods, including our proposed \oursjoint and \ourstile variants, alongside baseline methods such as Magnitude, Wanda, SparseGPT, Thanos, ProxSparse, and MaskLLM. These tables complement the average accuracy and perplexity results reported in Tables \ref{table::joint_patch} and \ref{table::tiled_patch} of the main paper, offering a granular view of model performance across individual tasks.

For smaller models (Qwen-2.5 0.5B, LLaMA-3.2 1B, and Gemma-3 1B), we report results using the \oursjoint variant, which jointly optimizes dense tile locations and sparsity patterns within sparse tiles. For larger models (LLaMA-2 7B and LLaMA-3.1 8B), we report results using the memory-efficient \ourstile variant, which optimizes dense tile selections with a fixed 2:4 sparsity mask. The per-task accuracies highlight the effectiveness of our approaches in maintaining robust performance across diverse tasks, even at high sparsity levels, compared to baseline methods.

The following tables detail the per-task accuracies for each model:

\begin{itemize}
    \item \textbf{Qwen-2.5 0.5B}: Table \ref{table:qwen_joint_patch} presents the per-task accuracies for the \oursjoint variant and baselines at 0\% and 50\% sparsity, with \oursjoint evaluated at 25\%, 35\%, and 45\% sparsity.
    \item \textbf{LLaMA-2 7B}: Table \ref{table:llama2_tile_patch} shows the per-task accuracies for the \ourstile variant and baselines, with \ourstile evaluated at 25\%, 35\%, and 45\% sparsity.
    \item \textbf{LLaMA-2 13B}: Table \ref{table:llama2_13b_tile_patch} shows the per-task accuracies for the \ourstile variant on LLaMA-2 13B at 25\%, 35\%, and 45\% sparsity, alongside Dense, Magnitude, Wanda, SparseGPT, Thanos, and ProxSparse baselines. MaskLLM is reported as N/A on LLaMA-2 13B because no public checkpoint is available and reproducing it would cost $\sim$2304 A100 GPU-hours.
    \item \textbf{LLaMA-3.1 8B}: Table \ref{table:llama31_tile_patch} provides the per-task accuracies for the \ourstile variant and baselines, with \ourstile at 25\%, 35\%, and 45\% sparsity.
    \item \textbf{LLaMA-3.2 1B}: Table \ref{table:llama32_joint_patch} reports the per-task accuracies for the \oursjoint variant and baselines, with \oursjoint at 25\%, 35\%, and 45\% sparsity.
    \item \textbf{Gemma-3 1B}: Table \ref{table:gemma_joint_patch} details the per-task accuracies for the \oursjoint variant and baselines, with \oursjoint at 25\%, 35\%, and 45\% sparsity.
\end{itemize}

\begin{revblock}
\paragraph{Where \ours degrades.} Degradation under pruning is task-dependent. Knowledge-intensive tasks suffer the largest drops and recover fastest as the sparsity budget is relaxed: on LLaMA-2 7B, MMLU falls from 41.82 (dense) to 27.28 at 45\% sparsity and returns to 32.33 at 25\%, the largest recovery of any task (Table~\ref{table:llama2_tile_patch}), while tasks such as PIQA, WinoGrande, and RACE stay within a few points of dense across all budgets. The allocation itself is not task-specific: \ours learns its masks from a task-agnostic language-modeling objective, so a single allocation serves all downstream tasks, and the learned allocations are stable across random seeds (Appendix~\ref{app::seed}) and calibration corpora (Appendix~\ref{app::c4_vs_slimpajama}), suggesting that they capture model-intrinsic redundancy rather than data- or task-specific structure. Learning task-conditioned allocations is an interesting direction for future work.
\end{revblock}

These results enable a deeper analysis of the task-specific performance trends, demonstrating the flexibility and robustness of \oursjoint and \ourstile in achieving high accuracy across diverse tasks while maintaining hardware-friendly sparsity patterns.

\begin{table}[H]
\centering
\small
\caption{Model quality (task accuracy across eight zero-shot tasks, reported in \%) for Qwen-2.5 0.5B with different pruning methods. \oursjoint optimizes dense tile locations and sparsity patterns, enabling a flexible sparsity-quality tradeoff.}
\label{table:qwen_joint_patch}
\setlength{\tabcolsep}{3pt}%
\resizebox{\textwidth}{!}{%
\begin{tabular}{l l l c c c c c c c c c}
\toprule
Sparsity & Method & Pattern & \taskname{MMLU} & \taskname{PIQA} & \taskname{ARC-E} & \taskname{ARC-C} & \taskname{WinoG.} & \taskname{OBQA} & \taskname{RACE} & \taskname{HellaS.} & \avgname{Avg} \\
\midrule
0\% & Dense & - & 47.71 & 70.24 & 64.48 & 29.52 & 56.20 & 24.20 & 35.02 & 40.63 & 46.00 \\
\midrule
50\% & Magnitude & 2:4 & 23.00 & 54.24 & 31.23 & 19.20 & 49.96 & 13.60 & 23.44 & 26.59 & 30.16 \\
     & Wanda & 2:4 & 24.43 & 58.71 & 43.18 & 17.75 & 51.62 & 12.20 & 26.32 & 29.58 & 32.97 \\
     & SparseGPT & 2:4 & 22.93 & 60.77 & 46.60 & 20.82 & 52.88 & 14.00 & 29.57 & 30.93 & 34.81 \\
     & Thanos & 2:4 & 22.97 & 60.17 & 45.37 & 19.20 & 53.59 & 15.20 & 31.00 & 31.31 & 34.85 \\
     & ProxSparse & 2:4 & 23.00 & 57.34 & 40.53 & 18.26 & 48.62 & 14.00 & 25.65 & 29.02 & 32.05 \\
     & MaskLLM & 2:4 & 25.11 & 67.03 & 56.57 & 23.98 & 52.57 & 20.20 & 33.30 & 35.90 & 39.33 \\
\midrule
\oursrow 45\% & \oursjoint & Dense/2:4 Tiles & 27.39 & 68.44 & 59.13 & 25.77 & 53.67 & 19.80 & 32.15 & 35.99 & 40.29 \\
\ourssep
\oursrow 35\% & \oursjoint & Dense/2:4 Tiles & 29.04 & 68.88 & 60.40 & 26.37 & 55.09 & 20.40 & 32.44 & 36.58 & 41.15 \\
\ourssep
\oursrow 25\% & \oursjoint & Dense/2:4 Tiles & 30.89 & 69.15 & 62.79 & 29.10 & 55.33 & 20.00 & 34.16 & 37.71 & 42.39 \\
\bottomrule
\end{tabular}
}

\end{table}

\begin{table}[H]
\centering
\small
\caption{Model quality (task accuracy across eight zero-shot tasks, reported in \%) for LLaMA-2 7B with different pruning methods. \ourstile optimizes tile-based sparsity, enabling a flexible sparsity-quality tradeoff.}
\label{table:llama2_tile_patch}
\setlength{\tabcolsep}{3pt}%
\resizebox{\textwidth}{!}{%
\begin{tabular}{l l l c c c c c c c c c}
\toprule
Sparsity & Method & Pattern & \taskname{MMLU} & \taskname{PIQA} & \taskname{ARC-E} & \taskname{ARC-C} & \taskname{WinoG.} & \taskname{OBQA} & \taskname{RACE} & \taskname{HellaS.} & \avgname{Avg} \\
\midrule
0\% & Dense & - & 41.82 & 78.07 & 76.35 & 43.52 & 69.06 & 31.40 & 39.52 & 57.13 & 54.61 \\
\midrule
50\% & Magnitude & 2:4 & 25.82 & 70.02 & 61.78 & 30.12 & 61.01 & 21.80 & 31.48 & 45.45 & 43.44 \\
     & Wanda & 2:4 & 25.80 & 71.00 & 63.80 & 30.29 & 61.09 & 25.20 & 35.50 & 41.75 & 44.30 \\
     & SparseGPT & 2:4 & 26.17 & 70.73 & 63.80 & 30.63 & 65.04 & 24.00 & 37.13 & 43.18 & 45.09 \\
     & Thanos & 2:4 & 25.27 & 70.78 & 63.43 & 30.97 & 64.56 & 23.80 & 36.46 & 43.11 & 44.80 \\
     & ProxSparse & 2:4 & 26.77 & 71.60 & 65.70 & 33.02 & 62.90 & 24.20 & 35.31 & 47.84 & 45.92 \\
     & MaskLLM & 2:4 & 27.65 & 74.76 & 69.44 & 35.58 & 65.04 & 26.80 & 38.56 & 51.15 & 48.62 \\
\midrule
\oursrow 45\% & \ourstile & Dense/2:4 Tiles & 27.28 & 75.41 & 70.16 & 35.84 & 65.27 & 27.60 & 38.76 & 51.61 & 48.99 \\
\ourssep
\oursrow 35\% & \ourstile & Dense/2:4 Tiles & 29.93 & 76.71 & 70.88 & 36.95 & 65.67 & 28.20 & 39.33 & 52.96 & 50.08 \\
\ourssep
\oursrow 25\% & \ourstile & Dense/2:4 Tiles & 32.33 & 76.99 & 72.81 & 38.57 & 68.27 & 29.80 & 39.52 & 54.34 & 51.58 \\
\bottomrule
\end{tabular}
}

\end{table}

\begin{table}[H]
\centering
\small
\caption{Model quality (task accuracy across eight zero-shot tasks, reported in \%) for LLaMA-2 13B with different pruning methods. \ourstile optimizes tile-based sparsity, enabling a flexible sparsity-quality tradeoff. The Avg column reports the eight-task average. MaskLLM is omitted from this table because no public LLaMA-2 13B checkpoint is available and reproducing it would require approximately 2304 A100 GPU-hours, beyond our compute budget.}
\label{table:llama2_13b_tile_patch}
\setlength{\tabcolsep}{3pt}%
\resizebox{\textwidth}{!}{%
\begin{tabular}{l l l c c c c c c c c c}
\toprule
Sparsity & Method & Pattern & \taskname{MMLU} & \taskname{PIQA} & \taskname{ARC-E} & \taskname{ARC-C} & \taskname{WinoG.} & \taskname{OBQA} & \taskname{RACE} & \taskname{HellaS.} & \avgname{Avg} \\
\midrule
0\% & Dense & - & 52.07 & 79.16 & 79.42 & 48.46 & 71.98 & 35.40 & 40.48 & 60.08 & 58.38 \\
\midrule
50\% & Magnitude  & 2:4 & 27.53 & 72.03 & 62.46 & 32.17 & 62.35 & 24.20 & 36.65 & 50.10 & 45.94 \\
     & Wanda      & 2:4 & 29.51 & 73.01 & 68.90 & 35.07 & 66.93 & 24.80 & 38.76 & 46.61 & 47.95 \\
     & SparseGPT  & 2:4 & 33.42 & 73.56 & 68.60 & 36.86 & 69.61 & 28.00 & 39.52 & 47.78 & 49.67 \\
     & Thanos     & 2:4 & 33.51 & 73.50 & 68.90 & 36.92 & 66.90 & 28.00 & 39.03 & 47.86 & 49.33 \\
     & ProxSparse & 2:4 & 34.86 & 75.68 & 71.46 & 38.31 & 66.85 & 28.60 & 37.51 & 53.09 & 50.80 \\
\midrule
\oursrow 45\% & \ourstile & Dense/2:4 Tiles & 41.67 & 76.88 & 73.02 & 40.44 & 70.24 & 30.40 & 38.09 & 55.16 & 53.24 \\
\ourssep
\oursrow 35\% & \ourstile & Dense/2:4 Tiles & 41.07 & 77.75 & 75.55 & 44.03 & 70.80 & 31.20 & 39.43 & 56.95 & 54.60 \\
\ourssep
\oursrow 25\% & \ourstile & Dense/2:4 Tiles & 47.24 & 78.13 & 76.81 & 45.73 & 71.19 & 34.20 & 38.37 & 58.76 & 56.31 \\
\bottomrule
\end{tabular}
}
\end{table}

\begin{table}[H]
\centering
\small
\caption{Model quality (task accuracy across eight zero-shot tasks, reported in \%) for LLaMA-3.1 8B with different pruning methods. \ourstile optimizes tile-based sparsity, enabling a flexible sparsity-quality tradeoff.}
\label{table:llama31_tile_patch}
\setlength{\tabcolsep}{3pt}%
\resizebox{\textwidth}{!}{%
\begin{tabular}{l l l c c c c c c c c c}
\toprule
Sparsity & Method & Pattern & \taskname{MMLU} & \taskname{PIQA} & \taskname{ARC-E} & \taskname{ARC-C} & \taskname{WinoG.} & \taskname{OBQA} & \taskname{RACE} & \taskname{HellaS.} & \avgname{Avg} \\
\midrule
0\% & Dense & - & 63.57 & 80.09 & 81.44 & 51.37 & 73.48 & 33.40 & 39.14 & 60.02 & 60.31 \\
\midrule
50\% & Magnitude & 2:4 & 23.06 & 63.82 & 45.33 & 25.94 & 53.91 & 15.20 & 26.70 & 33.49 & 35.93 \\
     & Wanda & 2:4 & 27.85 & 68.88 & 58.33 & 26.71 & 60.93 & 19.00 & 33.78 & 38.70 & 41.77 \\
     & SparseGPT & 2:4 & 31.82 & 70.46 & 63.85 & 31.74 & 64.56 & 21.60 & 37.22 & 42.99 & 45.53 \\
     & Thanos & 2:4 & 34.23 & 70.40 & 63.13 & 31.40 & 63.61 & 23.20 & 37.03 & 42.75 & 45.72 \\
     & ProxSparse & 2:4 & 29.89 & 71.71 & 62.63 & 33.28 & 58.56 & 23.80 & 35.22 & 46.03 & 45.14 \\
     & MaskLLM & 2:4 & 42.47 & 77.04 & 73.15 & 40.19 & 68.43 & 28.80 & 38.28 & 54.04 & 52.80 \\
\midrule
\oursrow 45\% & \ourstile & Dense/2:4 Tiles & 47.32 & 77.96 & 73.61 & 41.89 & 68.03 & 29.00 & 36.56 & 54.44 & 53.60 \\
\ourssep
\oursrow 35\% & \ourstile & Dense/2:4 Tiles & 51.15 & 77.97 & 76.14 & 42.41 & 69.46 & 31.40 & 38.18 & 55.54 & 55.28 \\
\ourssep
\oursrow 25\% & \ourstile & Dense/2:4 Tiles & 52.95 & 77.75 & 77.57 & 44.62 & 70.56 & 31.80 & 39.90 & 56.69 & 56.48 \\
\bottomrule
\end{tabular}
}

\end{table}

\begin{table}[H]
\centering
\small
\caption{Model quality (task accuracy across eight zero-shot tasks, reported in \%) for LLaMA-3.2 1B with different pruning methods. \oursjoint optimizes dense tile locations and sparsity patterns, enabling a flexible sparsity-quality tradeoff.}
\label{table:llama32_joint_patch}
\setlength{\tabcolsep}{3pt}%
\resizebox{\textwidth}{!}{%
\begin{tabular}{l l l c c c c c c c c c}
\toprule
Sparsity & Method & Pattern & \taskname{MMLU} & \taskname{PIQA} & \taskname{ARC-E} & \taskname{ARC-C} & \taskname{WinoG.} & \taskname{OBQA} & \taskname{RACE} & \taskname{HellaS.} & \avgname{Avg} \\
\midrule
0\% & Dense & - & 37.57 & 74.54 & 65.53 & 31.32 & 60.62 & 26.40 & 37.89 & 47.76 & 47.70 \\
\midrule
50\% & Magnitude & 2:4 & 23.31 & 53.81 & 27.74 & 18.94 & 51.38 & 11.80 & 24.02 & 26.26 & 29.66 \\
     & Wanda & 2:4 & 22.90 & 58.11 & 37.08 & 19.20 & 49.09 & 13.20 & 25.17 & 28.11 & 31.61 \\
     & SparseGPT & 2:4 & 22.93 & 61.43 & 45.03 & 22.35 & 54.93 & 15.80 & 29.86 & 32.08 & 35.55 \\
     & Thanos & 2:4 & 23.12 & 62.40 & 44.91 & 21.76 & 54.30 & 16.00 & 31.10 & 32.09 & 35.71 \\
     & ProxSparse & 2:4 & 22.96 & 60.83 & 39.44 & 20.31 & 51.54 & 16.80 & 25.17 & 31.37 & 33.55 \\
     & MaskLLM & 2:4 & 26.28 & 69.10 & 57.41 & 25.85 & 55.48 & 21.40 & 32.82 & 39.94 & 41.04 \\
\midrule
\oursrow 45\% & \oursjoint & Dense/2:4 Tiles & 23.81 & 70.89 & 60.77 & 27.22 & 56.27 & 22.80 & 34.07 & 40.78 & 42.08 \\
\ourssep
\oursrow 35\% & \oursjoint & Dense/2:4 Tiles & 25.13 & 71.32 & 60.27 & 29.18 & 57.06 & 22.00 & 34.64 & 42.17 & 42.72 \\
\ourssep
\oursrow 25\% & \oursjoint & Dense/2:4 Tiles & 28.59 & 71.44 & 61.57 & 28.67 & 58.25 & 23.20 & 35.22 & 43.52 & 43.81 \\
\bottomrule
\end{tabular}
}

\end{table}

\begin{table}[H]
\centering
\small
\caption{Model quality (accuracy across eight zero-shot tasks) for Gemma-3 1B with different pruning methods. \oursjoint optimizes dense tile locations and sparsity patterns, enabling a flexible sparsity-quality tradeoff.}
\label{table:gemma_joint_patch}
\setlength{\tabcolsep}{3pt}%
\resizebox{\textwidth}{!}{%
\begin{tabular}{l l l c c c c c c c c c}
\toprule
Sparsity & Method & Pattern & \taskname{MMLU} & \taskname{PIQA} & \taskname{ARC-E} & \taskname{ARC-C} & \taskname{WinoG.} & \taskname{OBQA} & \taskname{RACE} & \taskname{HellaS.} & \avgname{Avg} \\
\midrule
0\% & Dense & - & 24.95 & 75.03 & 71.84 & 34.90 & 58.64 & 28.60 & 34.83 & 47.26 & 47.01 \\
\midrule
50\% & Magnitude & 2:4 & 23.08 & 59.79 & 37.29 & 17.66 & 50.59 & 14.00 & 22.87 & 27.97 & 31.66 \\
     & Wanda & 2:4 & 23.96 & 59.52 & 48.02 & 18.34 & 51.22 & 14.20 & 27.85 & 30.18 & 34.16 \\
     & SparseGPT & 2:4 & 23.62 & 62.79 & 49.83 & 19.03 & 51.54 & 15.20 & 30.62 & 31.99 & 35.58 \\
     & Thanos & 2:4 & 23.44 & 62.24 & 48.86 & 18.34 & 50.12 & 15.60 & 30.81 & 31.28 & 35.09 \\
     & ProxSparse & 2:4 & 23.10 & 64.25 & 50.72 & 21.59 & 53.43 & 18.00 & 29.09 & 32.86 & 36.63 \\
     & MaskLLM & 2:4 & 25.03 & 69.91 & 60.27 & 27.65 & 56.27 & 21.20 & 34.55 & 39.84 & 41.84 \\
\midrule
\oursrow 45\% & \oursjoint & Dense/2:4 Tiles & 23.54 & 71.65 & 63.97 & 27.47 & 57.30 & 23.60 & 33.49 & 41.39 & 42.80 \\
\ourssep
\oursrow 35\% & \oursjoint & Dense/2:4 Tiles & 25.38 & 72.31 & 63.80 & 27.39 & 56.67 & 24.00 & 34.74 & 42.07 & 43.30 \\
\ourssep
\oursrow 25\% & \oursjoint & Dense/2:4 Tiles & 25.45 & 71.87 & 66.16 & 30.55 & 57.85 & 22.80 & 34.55 & 43.33 & 44.07 \\
\bottomrule
\end{tabular}
}

\end{table}

\section{Tile transfer learning}
\label{app::tile_transfer_learning}

We also test whether initializing tile logits with priors from one-shot pruning methods improves performance, as done in MaskLLM \citep{maskllm}. In our case, the initialization is derived from one-shot pruning with unstructured sparsity. We initialize tiles that retain more nonzeros after unstructured pruning with positive logits (favoring dense assignment), while the remaining tiles receive negative logits, controlled by a strength parameter. The number of tiles initialized as dense is selected such that the overall layer-wise sparsity target is satisfied. As shown in Table \ref{ablation:tile_prior}, the choice of prior has little impact on final performance: all priors yield nearly identical perplexity, with random initialization often performing best. This is likely because the global sparsity target enables dynamic reallocation of sparsity across layers during training, overriding the effect of any fixed initialization. For consistency with prior work, we adopt SparseGPT initialization in all experiments.

\begin{table}[H]
\centering
\small
\setlength{\tabcolsep}{6pt}
\begin{threeparttable}
\caption{Perplexity ($\downarrow$) under different tile prior initializations. All priors yield nearly identical performance, suggesting that the global sparsity target allows dynamic reallocation of sparsity during training, overriding the influence of fixed initialization.}

\label{ablation:tile_prior}
\begin{tabular}{l c c c c c}
\toprule
Sparsity (0.5B) & Nothing & SparseGPT & Wanda & Magnitude & Random \\
\midrule
45\% & 14.80 & 14.57 & 14.50 & \textbf{14.48} & 14.51 \\
35\% & 13.97 & 13.84 & 13.87 & 13.85 & \textbf{13.79} \\
25\% & 13.47 & 13.47 & 13.37 & 13.44 & \textbf{13.33} \\
\bottomrule
\end{tabular}
\end{threeparttable}
\end{table}

\section{Implementation details and hyperparameters}
\label{app::implementation_details}

We train all masks using the HuggingFace Trainer API \citep{huggingface} for 2000 steps with a global batch size of 256 and a sequence length of 4096, resulting in 2B tokens processed from the SlimPajama corpus \citep{slimpajama}. 

Training is accelerated via data parallelism across a single node with 4 H100 GPUs. In this setup, \oursjoint requires 18 and 24 GPU hours on the 0.5B and 1B models, respectively, while \ourstile requires 84, 96, and 160 GPU hours on the 7B, 8B, and 13B models, respectively (40 hours of wall-clock time for the 13B model).

The hyperparameters for \oursjoint and \ourstile are summarized in Table~\ref{table::hyper_params}, tuned on Qwen-2.5-0.5B. For the 2:4 mask parameters, we follow the configuration from MaskLLM \citep{maskllm}.  

\begin{table}[H]
    \centering
        \caption{Hyper-parameters used for \oursjoint and \ourstile across sparsity ratios. All hyper parameters were tuned on Qwen-2.5-0.5B.}
    \label{table::hyper_params}
    \resizebox{\linewidth}{!}{
    \begin{tabular}{c c c c c c c c c}
    \toprule
        Sparsity & Method & Optimizer & Logits Init &
        Gumbel Scaling &
        Gumbel &
        Prior(Strength)  &
        Sparse Reg. & Weight Reg. 
 \\
    \midrule
       25\% & \oursjoint  & Adam(0.001)  & $\mathcal{N}(0, 0.014)$ & $25 \rightarrow 350$  & $2 \rightarrow 0.05$ & SparseGPT($3$) & 7 & 10 \\
       35\% & \oursjoint  & Adam(0.001)  & $\mathcal{N}(0, 0.014)$ & $25 \rightarrow 350$  & $2 \rightarrow 0.05$ & SparseGPT($3$) & 7 & 10 \\
       45\% & \oursjoint  & Adam(0.001)  & $\mathcal{N}(0, 0.014)$ & $25 \rightarrow 350$  & $4 \rightarrow 0.05$ & SparseGPT($3$) & 7 & 10 \\
    \midrule
       25\% & \ourstile   & Adam(0.0001) & $\mathcal{N}(0, 0.014)$ & $100 \rightarrow 500$ & $2 \rightarrow 0.05$ & SparseGPT($3$) & 3 & 0.1 \\
       35\% & \ourstile   & Adam(0.0001) & $\mathcal{N}(0, 0.014)$ & $100 \rightarrow 500$ & $2 \rightarrow 0.05$ & SparseGPT($3$) & 3 & 0.1 \\
       45\% & \ourstile   & Adam(0.0001) & $\mathcal{N}(0, 0.014)$ & $100 \rightarrow 500$ & $2 \rightarrow 0.05$ & SparseGPT($3$) & 3 & 0.1 \\
    \bottomrule
    \end{tabular}
    }

\end{table}

\begin{revblock}
\paragraph{Implementation of \ourstile.} \ourstile requires only a minimal change to the joint training procedure. We first obtain a 2:4 mask, either with a one-shot pruning method such as SparseGPT or Wanda, or by training MaskLLM. The resulting binary (0/1) 2:4 mask is saved and kept fixed throughout training: Equation~\ref{eq:tile_merge} is unchanged except that $\Tilde{\mM}_{2:4}$ is a constant binary mask instead of a learnable differentiable one. The trainable parameters therefore reduce from both the tile and 2:4 mask parameters to a single scalar logit per tile, which substantially lowers memory use and training cost. Table~\ref{ablation:tile_only} compares the frozen-mask choices on Qwen-2.5 0.5B, where MaskLLM provides the strongest frozen mask. In all \ourstile experiments, the tile logits are initialized with the SparseGPT prior (Table~\ref{table::hyper_params}). For the results in Table~\ref{table::tiled_patch}, we freeze the publicly released MaskLLM masks for LLaMA-2 7B and LLaMA-3.1 8B; since no MaskLLM mask is available for LLaMA-2 13B, we freeze its SparseGPT 2:4 mask instead.
\end{revblock}

\section{Additional layer-wise sparsity distributions}
\label{app::sparsity_distribution}

\begin{figure}[htb]
\centering
\includegraphics[width=0.8\linewidth]{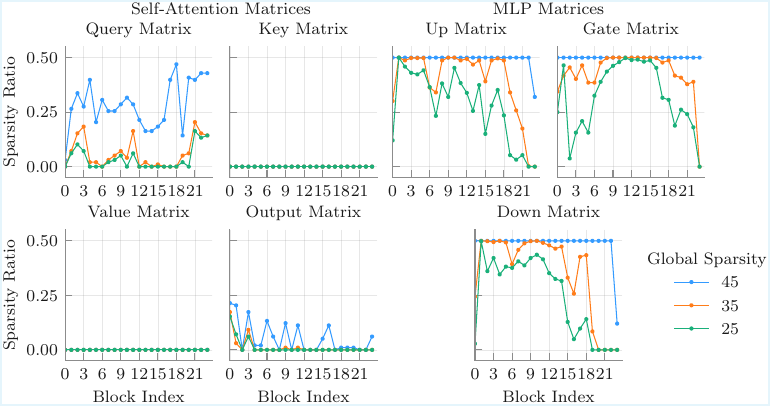}
\caption{Sparsity distribution across Attention and MLP layers under varying global sparsity budgets in Qwen-2.5 0.5B.}
\label{fig:linear_layer_sparsity_allocation_qwen}
\end{figure}

\begin{figure}[htb]
\begin{center}
\includegraphics[width=5.5in]{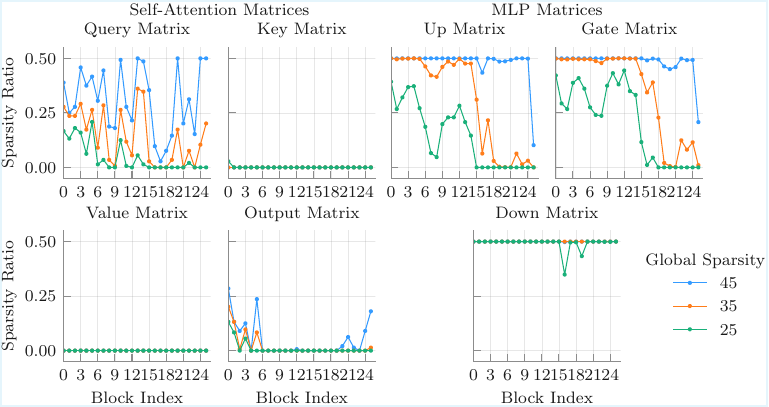}
\end{center}
\caption{Sparsity distribution across Attention and MLP layers under varying global sparsity budgets in Gemma-3 1B.}
\label{fig:linear_layer_sparsity_allocation_gemma}
\end{figure}

\begin{figure}[hbt]
\begin{center}
\includegraphics[width=5.5in]{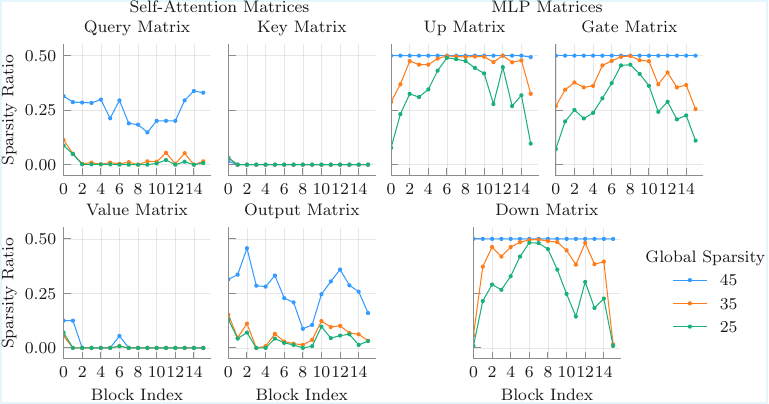}
\end{center}
\caption{Sparsity distribution across Attention and MLP layers under varying global sparsity budgets in LLaMA-3.2 1B.}
\label{fig:linear_layer_sparsity_allocation_llama}
\end{figure}

In this appendix, we provide the per-matrix sparsity distributions for \rev{Qwen-2.5 0.5B (Figure~\ref{fig:linear_layer_sparsity_allocation_qwen}, discussed in Section~\ref{sec:results}),} the Gemma-3-1B (Figure~\ref{fig:linear_layer_sparsity_allocation_gemma}) and Llama-3.2-1B (Figure~\ref{fig:linear_layer_sparsity_allocation_llama}) models, as referenced in the main text. Similar to the Qwen-2.5 0.5B model, the patterns observed here indicate that MLP layers (up, gate, and down matrices) are pruned more aggressively, absorbing the majority of sparsity. In contrast, the self-attention layers are treated as more critical, with key and value matrices remaining largely dense or unpruned, while the query matrix experiences the highest pruning within the attention submodule, and the output matrix shows moderate pruning under higher global sparsity targets. This consistent behavior across models underscores the redundancy in MLP components and the sensitivity of attention mechanisms.

\section{Comparison with unstructured sparsity}
\label{app::unstructured}

In this section, we compare the quality of the models pruned with \ours against other unstructured sparsity methods. Table \ref{table::unstructured} summarizes the average accuracy of the models across eight downstream tasks and the model perplexity on the WikiText2 dataset. The results indicate that while unstructured sparsity consistently outperforms the hybrid sparsity, the gap between the two is not significant, showing that \ours helps bridge the gap between unstructured sparsity and semi-structured sparsity.

\begin{table*}[ht]
\centering
\caption{Model quality (average accuracy across eight zero-shot tasks and perplexity on WikiText2 dataset) for \ours, Wanda, and SparseGPT. For models with less than or equal to 1B parameters, \oursjoint optimizes both dense tile locations and sparsity patterns, while for larger models \ourstile optimizes only dense tile locations with frozen sparsity patterns, both using Dense/2:4 Tiles pattern allowing continuous sparsity ratios and flexible tradeoffs between sparsity and model quality. Wanda and SparseGPT are unstructured pruning methods.}
\label{table::unstructured}
\small
\setlength{\tabcolsep}{3pt}
\resizebox{\textwidth}{!}{
\begin{tabular}{l l l c c c c c c c c c c}
\toprule
Sparsity & Method & Pattern &
\multicolumn{2}{c}{Qwen-2.5 0.5B} &
\multicolumn{2}{c}{LLaMA-3.2 1B} &
\multicolumn{2}{c}{Gemma-3 1B} &
\multicolumn{2}{c}{LLaMA-2 7B} &
\multicolumn{2}{c}{LLaMA-3.1 8B} \\
\cmidrule(lr){4-5} \cmidrule(lr){6-7} \cmidrule(lr){8-9} \cmidrule(lr){10-11} \cmidrule(lr){12-13}
& & & \taskname{Acc (\% $\uparrow$)} & \taskname{PPL ($\downarrow$)} & \taskname{Acc (\% $\uparrow$)} & \taskname{PPL ($\downarrow$)} & \taskname{Acc (\% $\uparrow$)} & \taskname{PPL ($\downarrow$)} & \taskname{Acc (\% $\uparrow$)} & \taskname{PPL ($\downarrow$)} & \taskname{Acc (\% $\uparrow$)} & \taskname{PPL ($\downarrow$)} \\
\midrule
\oursrow 45\% & \ours & Dense/2:4 Tiles & 40.29 & 14.57 & 42.08 & 12.23 & 42.80 & 11.96 & 48.99 & 6.55 & 53.60 & 8.20 \\
45\% & Wanda & Unstructured & 41.45 & 18.81 & 40.76 & 16.56 & 42.87 & 25.38 & 52.72 & 6.36 & 55.67 & 8.24 \\
45\% & SparseGPT & Unstructured & 42.31 & 17.65 & 42.66 & 15.01 & 43.52 & 22.26 & 52.77 & 6.46 & 56.70 & 8.21 \\
\midrule
\oursrow 35\% & \ours & Dense/2:4 Tiles & 41.15 & 13.84 & 42.72 & 11.67 & 43.30 & 11.48 & 50.08 & 6.18 & 55.28 & 7.89 \\
35\% & Wanda & Unstructured & 43.46 & 15.04 & 44.60 & 11.95 & 45.50 & 16.98 & 54.37 & 5.87 & 58.68 & 7.02 \\
35\% & SparseGPT & Unstructured & 44.66 & 14.79 & 45.62 & 11.68 & 45.45 & 16.92 & 54.18 & 5.92 & 58.81 & 7.07 \\
\midrule
\oursrow 25\% & \ours & Dense/2:4 Tiles & 42.39 & 13.47 & 43.81 & 11.00 & 44.07 & 11.17 & 51.58 & 5.86 & 56.48 & 7.34 \\
25\% & Wanda & Unstructured & 45.70 & 13.70 & 46.50 & 10.46 & 46.56 & 15.14 & 54.60 & 5.65 & 59.80 & 6.54 \\
25\% & SparseGPT & Unstructured & 45.28 & 13.63 & 46.52 & 10.42 & 46.37 & 15.05 & 54.71 & 5.68 & 59.52 & 6.55 \\
\bottomrule
\end{tabular}
}

\end{table*}

\section{Layer-wise sparsity distribution comparison with other work}
\label{app::comparison_sparsity}

We compare the sparsity allocation learned by \ours with OWL \citep{owl} and AlphaPruning \citep{alphapruning}. To ensure a robust baseline, we performed extensive hyperparameter sweeps for both methods and selected the configurations that achieved the best perplexity when applying Wanda as the underlying pruning operator:

\begin{itemize}
    \item OWL: We swept $\lambda \in \{0.01, 0.03, 0.05, 0.08, 0.1\}$ and $M \in \{3, 5, 7, 10\}$.
    \item AlphaPruning: We swept 20 values of the temperature parameter $\tau$ between 0 and 0.5.
\end{itemize}

A critical distinction is that OWL and AlphaPruning are primarily designed for unstructured sparsity. While OWL includes an N:M structured variant, it is restricted to a fixed 50\% global sparsity. Furthermore, OWL allocates sparsity at the block level (assigning uniform sparsity to all matrices within a Transformer block). Although the authors propose a weight-wise variant, they report, and our experiments confirm, that it yields inferior performance. In contrast, PATCH operates with tile-level granularity within a semi-structured constraint, offering a unique combination of hardware acceleration and fine-grained control.

\begin{figure}[htb]
\begin{center}
\includegraphics[width=5.5in]{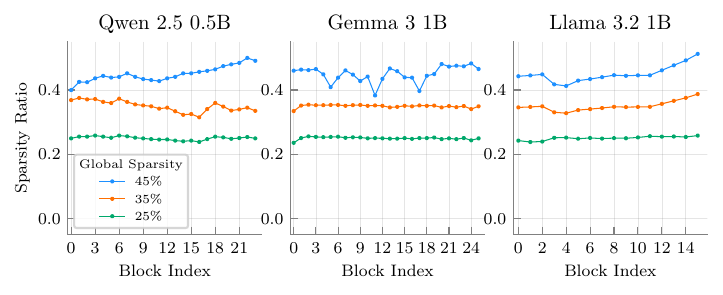}
\end{center}
\caption{Layer-wise sparsity distribution of OWL across models and global sparsity budgets.}
\label{fig:layerwise_owl}
\end{figure}

\begin{figure}[htb]
\begin{center}
\includegraphics[width=5.5in]{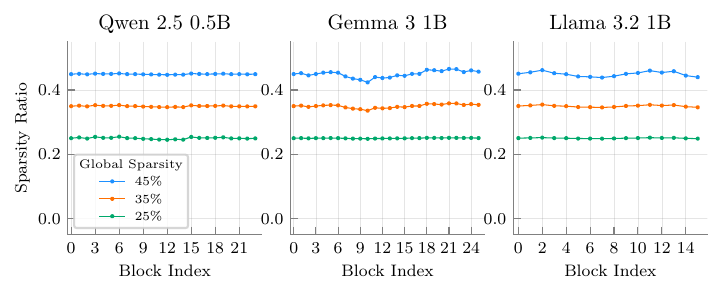}
\end{center}
\caption{Layer-wise sparsity distribution of AlphaPruning across models and global sparsity budgets.}
\label{fig:layerwise_alpha}
\end{figure}

As seen in Figures~\ref{fig:layerwise_owl} and \ref{fig:layerwise_alpha}, both baselines produce relatively flat sparsity distributions across the model. AlphaPruning exhibits only minor fluctuations in the middle layers, whereas PATCH discovers distinct, highly non-uniform patterns (e.g., preserving Attention layers while aggressively pruning MLP blocks).

\begin{figure}[htb]
\begin{center}
\includegraphics[width=5.5in]{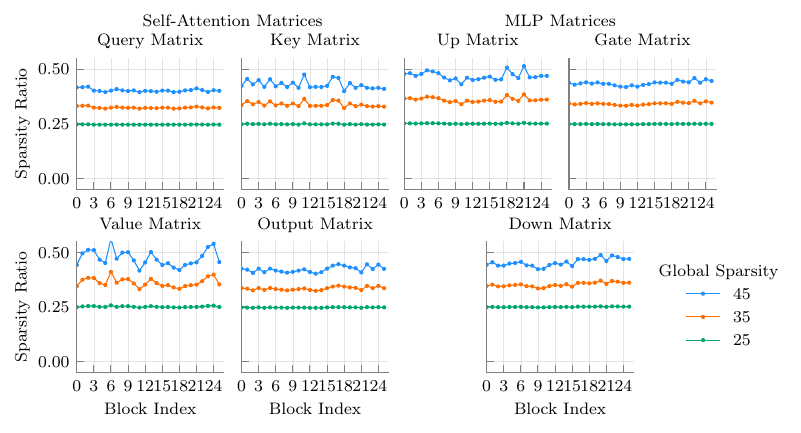}
\end{center}
\caption{AlphaPruning sparsity distribution across attention and MLP layers under varying global sparsity budgets in Gemma-3 1B.}
\label{fig:weightwise_alpha}
\end{figure}

For OWL, we performed a sweep over the $\lambda$ parameter in ${0.01, 0.03, 0.05, 0.08, 0.1}$ and the $M$ parameter in ${3, 5, 7, 10}$ for each model. For AlphaPruning, we swept 20 values of  $\tau \in [0, 0.5]$. For both methods, we selected the hyperparameters that achieved the best perplexity using Wanda as the pruning metric.

\section{Variation across seeds}
\label{app::seed}

We evaluate PATCH on Qwen-2.5 0.5B and Llama-3.2 1B across different seeds. As shown in Table~\ref{table:seed_ppl}, model performance remains consistent across seeds.

\begin{table}[h]
\centering
\caption{Perplexity across seeds for Qwen-2.5 0.5B and Llama-3.2 1B.}
\label{table:seed_ppl}
\begin{tabular}{lcccccc}
\toprule
Model & Sparsity (\%) & Seed 0 (default) & Seed 25 & Seed 26 & Seed 42 \\
\midrule
Qwen-2.5 0.5B & 25 & 13.47 & 13.41 & 13.38 & 13.36 \\
              & 35 & 13.84 & 13.89 & 13.85 & 13.84 \\
              & 45 & 14.57 & 14.59 & 14.61 & 14.49 \\
\midrule
Llama-3.2 1B  & 25 & 11.00 & 11.09 & 11.03 & 11.21 \\
              & 35 & 11.67 & 11.72 & 11.56 & 11.86 \\
              & 45 & 12.23 & 12.32 & 12.26 & 12.55 \\
\bottomrule
\end{tabular}
\end{table}

\begin{figure}[htb]
\begin{center}
\includegraphics[width=5.5in]{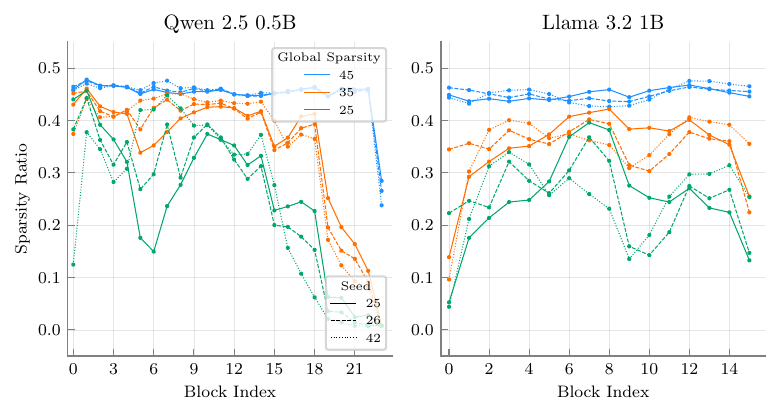}
\end{center}
\caption{Layer-wise sparsity distribution of PATCH across seeds.}
\label{fig:layerwise_seed}
\end{figure}

\begin{figure}[htb]
\begin{center}
\includegraphics[width=5.5in]{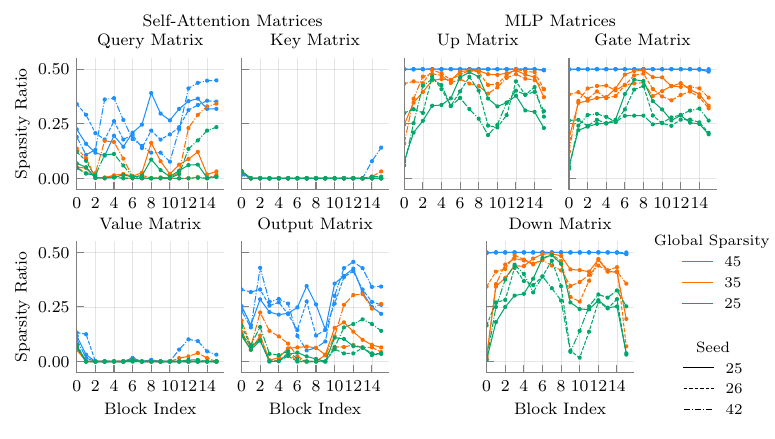}
\end{center}
\caption{Sparsity distribution across attention and MLP layers under varying global sparsity budgets in Llama-3.2 1B across seeds.}
\label{fig:weightwise_llama_seed}
\end{figure}

\begin{figure}[htb]
\begin{center}
\includegraphics[width=5.5in]{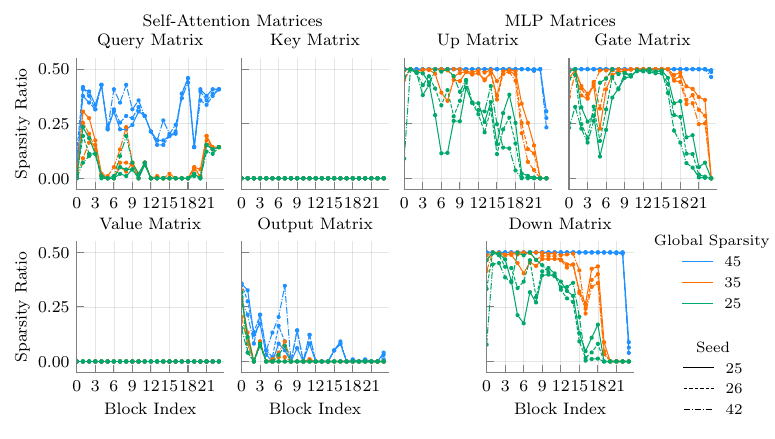}
\end{center}
\caption{Sparsity distribution across attention and MLP layers under varying global sparsity budgets in Qwen-2.5 0.5B across seeds.}
\label{fig:weightwise_qwen_seed}
\end{figure}

In addition, we include the corresponding sparsity allocations across layers (Figure~\ref{fig:layerwise_seed}) and across individual weight matrices for both Llama-3.2 1B (Figure~\ref{fig:weightwise_llama_seed}) and Qwen-2.5 0.5B (Figure~\ref{fig:weightwise_qwen_seed}). From these, we observe the following:

\begin{itemize}
    \item At the global block level,  the optimization consistently identifies the middle Transformer blocks as the most redundant (receiving the highest sparsity), while the initial and (especially) final blocks are pruned less.
    \item At the weight-matrix level, the allocation of sparsity between attention and MLP modules remains stable. For example, in Llama-3.2 1B (Figure \ref{fig:weightwise_llama_seed}), the Key and Value matrices consistently remain dense across all seeds, while the Query and MLP matrices absorb the majority of the sparsity.
\end{itemize}

While the specific tile indices may vary slightly due to the stochastic sampling, the macroscopic pruning strategy learned by PATCH is highly reproducible and robust to initialization.

\section{Fine-tuning after mask training}
\label{app::finetuning}

In this section, we present results from fine-tuning the remaining unpruned weights after mask training. We conducted a brief 5.6M-token fine-tuning run on the SlimPajama dataset\rev{, a small fraction of the 2B-token mask-search budget (Appendix~\ref{app::implementation_details})}.

Fine-tuning yields a consistent improvement in average zero-shot accuracy (e.g., +0.8\% for Llama-3.2 1B at 25\% sparsity and +0.6\% for Qwen-2.5 0.5B at 45\% sparsity), as seen in Table \ref{table:ft}. Interestingly, we observe a slight degradation in perplexity. We attribute this to the limited calibration data (5.6M tokens) compared to the trillions of tokens seen during pre-training; short fine-tuning can sometimes slightly drift the language modeling distribution while sharpening downstream task performance.

These results confirm that PATCH creates a high-quality sparsity topology that serves as a strong foundation. While the mask alone delivers state-of-the-art performance, subsequent fine-tuning, even with a limited budget, can further recover accuracy, offering a flexible path for users with additional compute resources.

\begin{table}[h!]
\centering
\caption{Model quality (average accuracy across eight zero-shot tasks and perplexity on WikiText2
dataset) for PATCH after a short fine-tuning.}
\label{table:ft}
\begin{tabular}{l c l c c}
\toprule
Model & Sparsity & Method & Wiki PPL ($\downarrow$) & Avg Acc (\% $\uparrow$) \\
\midrule
\textbf{Qwen-2.5 0.5B} & 45\% & \oursjoint & \textbf{14.57} & 40.29 \\
 &  & \oursjoint + FT & 14.96 & \textbf{40.87} \\
 & 35\% & \oursjoint & \textbf{13.84} & 41.15 \\
 &  & \oursjoint + FT & 14.32 & \textbf{41.59} \\
 & 25\% & \oursjoint & \textbf{13.47} & 42.39 \\
 &  & \oursjoint + FT & 13.85 & \textbf{42.55} \\
\midrule
\textbf{LLaMA-3.2 1B} & 45\% & \oursjoint & \textbf{12.23} & 42.08 \\
 &  & \oursjoint + FT & 12.64 & \textbf{42.94} \\
 & 35\% & \oursjoint & \textbf{11.67} & 42.72 \\
 &  & \oursjoint + FT & 12.02 & \textbf{43.50} \\
 & 25\% & \oursjoint & \textbf{11.00} & 43.81 \\
 &  & \oursjoint + FT & 11.36 & \textbf{44.61} \\
\bottomrule
\end{tabular}

\end{table}

\section{Comparison with FFN-only 2:4 pruning}
\label{app::ffn_only}

The sparsity allocation discovered by \ours places most of the sparsity inside FFN modules while keeping attention layers comparatively dense (Section~\ref{sec:results} and Appendix~\ref{app::sparsity_distribution}). A natural question is whether a simpler heuristic that prunes only the FFN layers to 2:4 sparsity, optionally leaving the last few decoder blocks dense to hit a target global sparsity, can match \our's accuracy while preserving its throughput advantage.

We construct such a baseline using SparseGPT to prune all FFN modules to 2:4 sparsity. To match \our's lower global sparsity targets ($\leq 35\%$), we keep the last 3--5 decoder blocks dense (denoted ``except last $k$''). Table~\ref{table::ffn_only_acc} reports accuracy and perplexity, and Table~\ref{table::ffn_only_speedup} reports throughput on an A6000 GPU under the same inference setup as Section~\ref{app::stoicc}.

At matched global sparsity, the FFN-only heuristic achieves throughput that is essentially identical to \ours (Table~\ref{table::ffn_only_speedup}), confirming that \our's tile-level heterogeneity introduces no measurable runtime overhead. However, \ours significantly outperforms the heuristic in accuracy across all evaluated models and sparsity levels (Table~\ref{table::ffn_only_acc}). For example, on LLaMA-2 7B at 45\% sparsity, \ours improves perplexity from 8.36 to 6.55 over an FFN-only baseline at lower global sparsity. This demonstrates that \our's learned, non-uniform tile assignments preserve critical FFN tiles and identify redundant attention tiles where appropriate, yielding a strictly better accuracy--efficiency trade-off than the rigid heuristic at the same hardware cost.

\begin{revblock}
\paragraph{Learned 2:4 masks with a heuristic allocation.} To test whether a stronger 2:4 mask closes the gap, we apply the same heuristic allocation on top of MaskLLM's learned 2:4 masks for Qwen-2.5 0.5B, LLaMA-3.2 1B, and Gemma-3 1B (Table~\ref{table::ffn_only_acc}). MaskLLM-based heuristics are far stronger than SparseGPT-based ones and occasionally match \our's perplexity, but they still reach lower average accuracy than \ours at comparable (and even lower) sparsity in every case, e.g., 39.34\% versus 40.29\% on Qwen-2.5 0.5B at 44\% versus 45\% sparsity. It is therefore the learned tile allocation, rather than simply reducing sparsity, that yields the accuracy gains; since \ours also substantially outperforms heuristics built on one-shot masks, both end-to-end optimization and adaptive tile allocation are necessary.
\end{revblock}

\begin{table}[h!]
\centering
\caption{Comparison of \ours against an FFN-only 2:4 pruning baseline at matched global sparsity. The baseline applies SparseGPT 2:4 pruning to FFN modules only, optionally keeping the last few decoder blocks dense (denoted ``except last $k$'') to reach the target global sparsity. \rev{Rows labeled MaskLLM apply MaskLLM's learned 2:4 masks to the FFNs with the same heuristic allocation.} \ours \rev{achieves higher average accuracy than every heuristic baseline at comparable sparsity, and lower perplexity in all but three cases, where a MaskLLM-based heuristic is marginally lower. Best value per model and sparsity group in \textbf{bold}.}}
\label{table::ffn_only_acc}
\setlength{\tabcolsep}{6pt}
\small
\begin{tabular}{l l c c c}
\toprule
Model & Method & Sparsity (\%) & PPL ($\downarrow$) & Avg Acc (\% $\uparrow$) \\
\midrule
\multirow{6}{*}{\textbf{Qwen-2.5 0.5B}}
 & \ours                          & 45 & 14.57 & \textbf{40.29} \\
 & SparseGPT (FFN only)           & 44 & 24.79 & 36.84 \\
 & \rev{MaskLLM (FFN only)}        & \rev{44} & \rev{\textbf{14.54}} & \rev{39.34} \\
 & \ours                          & 35 & \textbf{13.84} & \textbf{41.15} \\
 & SparseGPT (FFN except last 5)  & 35 & 20.46 & 38.33 \\
 & \rev{MaskLLM (FFN except last 5)} & \rev{35} & \rev{13.92} & \rev{40.59} \\
\midrule
\multirow{6}{*}{\textbf{LLaMA-3.2 1B}}
 & \ours                          & 45 & 12.23 & \textbf{42.08} \\
 & SparseGPT (FFN only)           & 41 & 19.97 & 38.73 \\
 & \rev{MaskLLM (FFN only)}        & \rev{41} & \rev{\textbf{12.13}} & \rev{41.36} \\
 & \ours                          & 35 & 11.67 & \textbf{42.72} \\
 & SparseGPT (FFN except last 3)  & 34 & 17.62 & \rev{39.43} \\
 & \rev{MaskLLM (FFN except last 3)} & \rev{34} & \rev{\textbf{11.39}} & \rev{42.45} \\
\midrule
\multirow{6}{*}{\textbf{Gemma-3 1B}}
 & \ours                          & 45 & \textbf{11.96} & \textbf{42.80} \\
 & SparseGPT (FFN only)           & 45 & 32.20 & 39.58 \\
 & \rev{MaskLLM (FFN only)}        & \rev{45} & \rev{12.10} & \rev{42.39} \\
 & \ours                          & 35 & \textbf{11.48} & \textbf{43.30} \\
 & SparseGPT (FFN except last \rev{5})  & 36 & 26.94 & 39.98 \\
 & \rev{MaskLLM (FFN except last 5)} & \rev{36} & \rev{11.6} & \rev{42.51} \\
\midrule
\multirow{3}{*}{\textbf{LLaMA-2 7B}}
 & \ours                          & 45 & \textbf{6.55} & 48.99 \\
 & SparseGPT (FFN only)           & 33 & 8.36 & 49.01 \\
 & \ours                          & 35 & \textbf{6.18} & \textbf{50.08} \\
\midrule
\multirow{4}{*}{\textbf{LLaMA-3.1 8B}}
 & \ours                          & 45 & \textbf{8.20} & \textbf{53.60} \\
 & SparseGPT (FFN only)           & 40 & 11.37 & 51.04 \\
 & \ours                          & 35 & \textbf{7.89} & \textbf{55.28} \\
 & SparseGPT (FFN except last 4)  & 35 & 10.30 & 51.49 \\
\bottomrule
\end{tabular}
\end{table}

\begin{table}[h!]
\centering
\caption{Throughput comparison of \ours against FFN-only 2:4 pruning under the same STOICC backend on an A6000 GPU (batch size 16, prefill length 128, generation length 128). At matched global sparsity, the FFN-only baseline matches \our's throughput, but \ours achieves significantly higher accuracy (Table~\ref{table::ffn_only_acc}). For reference, we also report fully 2:4 sparse models, which represent the upper bound of attainable acceleration.}
\label{table::ffn_only_speedup}
\setlength{\tabcolsep}{4pt}
\resizebox{\textwidth}{!}{
\begin{tabular}{l l c l c c}
\toprule
\textbf{Model} & \textbf{Method} & \textbf{Sparsity (\%)} & \textbf{Backend} & \textbf{Throughput (tok/s)} & \textbf{Speedup} \\
\midrule
\multirow{8}{*}{\textbf{LLaMA-2 7B}}
 & Dense                          & 0  & cuBLAS      & 1023.80 & 1.00$\times$ \\
 & \ours                          & 25 & STOICC      & 1212.79 & 1.18$\times$ \\
 & \ours                          & 35 & STOICC      & 1304.46 & 1.27$\times$ \\
 & \ours                          & 45 & STOICC      & 1410.20 & 1.38$\times$ \\
 & SparseGPT (FFN only)           & 33 & cuSPARSELt  & 1276.67 & 1.25$\times$ \\
 & SparseGPT (FFN only)           & 33 & STOICC      & 1302.27 & 1.27$\times$ \\
 & SparseGPT (full 2:4)           & 50 & cuSPARSELt  & 1434.25 & 1.40$\times$ \\
 & SparseGPT (full 2:4)           & 50 & STOICC      & 1501.91 & 1.47$\times$ \\
\midrule
\multirow{9}{*}{\textbf{LLaMA-3.1 8B}}
 & Dense                          & 0  & cuBLAS      & 908.02  & 1.00$\times$ \\
 & \ours                          & 25 & STOICC      & 1027.44 & 1.13$\times$ \\
 & \ours                          & 35 & STOICC      & 1079.02 & 1.19$\times$ \\
 & \ours                          & 45 & STOICC      & 1164.57 & 1.28$\times$ \\
 & SparseGPT (FFN except last 4)  & 35 & cuSPARSELt  & 1086.91 & 1.20$\times$ \\
 & SparseGPT (FFN except last 4)  & 35 & STOICC      & 1093.27 & 1.20$\times$ \\
 & SparseGPT (FFN only)           & 40 & cuSPARSELt  & 1122.31 & 1.24$\times$ \\
 & SparseGPT (FFN only)           & 40 & STOICC      & 1130.60 & 1.25$\times$ \\
 & SparseGPT (full 2:4)           & 50 & cuSPARSELt  & 1176.45 & 1.30$\times$ \\
 & SparseGPT (full 2:4)           & 50 & STOICC      & 1226.36 & 1.35$\times$ \\
\bottomrule
\end{tabular}
}
\end{table}

\section{Sparsity-matched comparison with SparseGPT 3:4}
\label{app::sparsegpt_3_4}

\ours at 25\% global sparsity and SparseGPT pruned with a 3:4 pattern both result in 75\% global density, allowing a direct comparison at a matched compression ratio. The 3:4 pattern is more flexible than the 2:4 pattern used inside \our's sparse tiles (it removes any one weight in every group of four rather than two), and as a result SparseGPT 3:4 matches or slightly outperforms \ours at 25\% in perplexity and average accuracy on most evaluated models (Table~\ref{table::sparsegpt_3_4}).

The crucial difference is hardware compatibility. NVIDIA Sparse Tensor Cores require exactly two non-zeros per group of four, so the 3:4 pattern cannot be accelerated and must run at dense speeds. \ours instead mixes dense tiles with hardware-supported 2:4 sparse tiles, which combined with the STOICC backend delivers $\sim$1.18$\times$ end-to-end speedup on commodity GPUs at the same compression ratio. This positions \ours as the practically-useful operating point at 75\% density: a small accuracy concession in exchange for measurable wall-clock acceleration.

\begin{table}[h!]
\centering
\caption{Sparsity-matched comparison between \ours at 25\% sparsity and SparseGPT pruned to a 3:4 pattern (75\% density). While SparseGPT 3:4 achieves slightly lower perplexity due to its higher per-group flexibility, the 3:4 pattern is incompatible with NVIDIA Sparse Tensor Cores (which require exactly two non-zeros per group of four) and therefore cannot be accelerated, running at dense speeds. \ours at 25\% sparsity, in contrast, mixes dense and 2:4 tiles and unlocks $\sim$1.18$\times$ end-to-end speedup on standard GPUs.}
\label{table::sparsegpt_3_4}
\setlength{\tabcolsep}{6pt}
\begin{tabular}{l l c c c}
\toprule
Model & Method & Sparsity (\%) & PPL ($\downarrow$) & Avg Acc (\% $\uparrow$) \\
\midrule
\multirow{2}{*}{\textbf{Qwen-2.5 0.5B}}
 & SparseGPT (3:4) & 25 & 13.99 & \textbf{43.34} \\
 & \ours           & 25 & \textbf{13.47} & 42.39 \\
\midrule
\multirow{2}{*}{\textbf{LLaMA-3.2 1B}}
 & SparseGPT (3:4) & 25 & \textbf{10.88} & \textbf{44.66} \\
 & \ours           & 25 & 11.00 & 43.81 \\
\midrule
\multirow{2}{*}{\textbf{Gemma-3 1B}}
 & SparseGPT (3:4) & 25 & 13.96 & \textbf{45.27} \\
 & \ours           & 25 & \textbf{11.17} & 44.07 \\
\midrule
\multirow{2}{*}{\textbf{LLaMA-2 7B}}
 & SparseGPT (3:4) & 25 & \textbf{5.55} & \textbf{54.08} \\
 & \ours           & 25 & 5.86 & 51.58 \\
\midrule
\multirow{2}{*}{\textbf{LLaMA-3.1 8B}}
 & SparseGPT (3:4) & 25 & \textbf{6.70} & \textbf{58.14} \\
 & \ours           & 25 & 7.34 & 56.48 \\
\bottomrule
\end{tabular}
\end{table}

\section{Robustness across calibration datasets}
\label{app::c4_vs_slimpajama}

To assess how sensitive \ours is to the choice of calibration corpus during mask training, we re-train Qwen-2.5 0.5B and LLaMA-3.2 1B with masks learned on the C4 dataset rather than SlimPajama, keeping all other hyperparameters identical. We compare against MaskLLM trained under the same conditions.

As shown in Table~\ref{table::c4_vs_slimpajama}, \ours produces consistent results across both calibration sources. Perplexity on WikiText2 is marginally higher when training on C4, which is expected given the domain shift between the calibration set and the evaluation distribution, but downstream zero-shot accuracy stays within roughly 0.6\% of the SlimPajama baseline. \ours also remains stronger than MaskLLM under either calibration corpus, indicating that the learned, non-uniform sparsity allocation is robust to the choice of training data.

\begin{table}[h!]
\centering
\caption{Robustness of \ours mask training to the choice of calibration corpus. We train identical \ours configurations on the C4 dataset and on SlimPajama and report WikiText2 perplexity and average zero-shot accuracy. Despite the domain shift, downstream accuracy remains within $\sim$0.6\% across datasets, and \ours outperforms MaskLLM under both calibration corpora.}
\label{table::c4_vs_slimpajama}
\setlength{\tabcolsep}{4pt}
\resizebox{\textwidth}{!}{
\begin{tabular}{l l c c c c}
\toprule
Model & Method & C4 PPL ($\downarrow$) & C4 Avg Acc (\% $\uparrow$) & SlimPajama PPL ($\downarrow$) & SlimPajama Avg Acc (\% $\uparrow$) \\
\midrule
\multirow{4}{*}{\textbf{Qwen-2.5 0.5B}}
 & MaskLLM (50\%)  & 16.27 & 38.78 & 15.22 & 39.33 \\
 & \ours (45\%)    & 15.30 & 40.02 & 14.57 & 40.29 \\
 & \ours (35\%)    & 14.34 & 40.94 & 13.84 & 41.15 \\
 & \ours (25\%)    & 13.81 & 41.91 & 13.47 & 42.39 \\
\midrule
\multirow{4}{*}{\textbf{LLaMA-3.2 1B}}
 & MaskLLM (50\%)  & 14.23 & 40.97 & 12.93 & 41.04 \\
 & \ours (45\%)    & 13.10 & 41.73 & 12.23 & 42.08 \\
 & \ours (35\%)    & 12.07 & 42.98 & 11.67 & 42.72 \\
 & \ours (25\%)    & 11.42 & 43.37 & 11.00 & 43.81 \\
\bottomrule
\end{tabular}
}
\end{table}

\section{Tile size and context length effects on speedup}
\label{app::tile_speedup}

\paragraph{Execution tile size.} The \rev{mask} tile size in \ours defines the granularity of the sparsity search space. The actual execution tile size is selected by STOICC's autotuner, which benchmarks candidate kernel configurations and picks the fastest one for each layer. To probe how the choice of execution tile interacts with throughput, we restrict the autotuner to a maximum tile size of either 64$\times$64 or 32$\times$64; smaller tiles than this are not supported by STOICC's metadata layout and compression constraints, and Triton additionally enforces a minimum block size of $K \geq 16$ for tensor-core dot products.

As shown in Table~\ref{table::tile_speedup}, capping execution tiles at 64$\times$64 has only a small effect on throughput: STOICC's unconstrained autotuner already prefers 64$\times$64 in most layers of our main experiments. Forcing 32$\times$64, by contrast, markedly slows down the kernels because tiles this small underutilize the tensor cores. Combined with Table~\ref{ablation:tile_size}, which shows that \rev{mask} tile size has minimal effect on accuracy, this analysis confirms that tile granularity is principally a hardware-efficiency knob rather than a model-quality knob, and that 128$\times$128 mask tiles, which the compiler subdivides into 64$\times$64 execution tiles, give the best end-to-end throughput.

\begin{table}[h!]
\centering
\caption{Effect of restricting STOICC's autotuner to specific maximum execution tile sizes on \our's end-to-end throughput, measured on LLaMA-2 7B with an A6000 GPU (batch size 16, prefill length 128, generation length 128). Limiting the autotuner to 64$\times$64 has a small effect on throughput (the unrestricted autotuner already prefers this size in most layers), while restricting it to 32$\times$64 underutilizes Tensor Cores and yields slower kernels. Tiles smaller than 32$\times$64 are not supported by STOICC's metadata layout and compression constraints. As reported in Table~\ref{ablation:tile_size}, model accuracy is largely insensitive to tile size, so the chosen 128$\times$128 mask-tile granularity (which the compiler subdivides at execution) provides the best balance between accuracy and hardware efficiency.}
\label{table::tile_speedup}
\setlength{\tabcolsep}{8pt}
\begin{tabular}{c c c c}
\toprule
Sparsity (\%) & Max.\ Execution Tile Size & Throughput (tok/s) & Speedup vs. Dense \\
\midrule
0  & ---              & 1023.80          & 1.00$\times$ \\
\midrule
\multirow{2}{*}{25}
   & 64$\times$64     & 1207.08          & 1.18$\times$ \\
   & 32$\times$64     & 1075.98          & 1.05$\times$ \\
\midrule
\multirow{2}{*}{35}
   & 64$\times$64     & 1288.54          & 1.26$\times$ \\
   & 32$\times$64     & 1142.65          & 1.12$\times$ \\
\midrule
\multirow{2}{*}{45}
   & 64$\times$64     & 1382.13          & 1.35$\times$ \\
   & 32$\times$64     & 1166.19          & 1.14$\times$ \\
\bottomrule
\end{tabular}
\end{table}

\paragraph{Long-context inference.} We also report \our's end-to-end speedup at longer prefill contexts. Table~\ref{table::long_context} shows speedups for LLaMA-2 7B at prefill lengths of 8K and 16K\rev{, alongside the 128-token prefill of Table~\ref{table::speedup}, with generation lengths of 128 and 1024}. As the prefill grows, attention computation increasingly dominates runtime, which reduces the relative impact of accelerating the FFN GEMMs that \ours targets. Speedup at 45\% sparsity drops from 1.38$\times$ at the 128-token prefill (Table~\ref{table::speedup}) to 1.22$\times$ at 8K and 1.14$\times$ at 16K. \rev{This behavior is not specific to \ours: attention cost grows quadratically with sequence length, whereas the GEMMs accelerated by weight sparsity (like other weight-compression methods, such as weight quantization) scale linearly. To reduce the attention overhead, we additionally benchmark \ours with the FlashAttention-2 (FA2) backend~\citep{flashattention2}, measuring each speedup relative to the dense model with the same backend. FA2 consistently increases \our's speedup, by up to 0.13$\times$ (128-token prefill, 1024 generated tokens, 45\% sparsity, approximately 10\% relative), because a larger fraction of inference time is then spent in the sparse GEMMs; with or without FA2, \ours retains a speedup at every sparsity level, including 25\%. The improvement is most pronounced when decoding longer sequences, where the savings from the repeated sparse linear layers are amortized over many decoding steps.}

\begin{table}[h!]
\centering
\caption{End-to-end throughput speedup of \ours over a dense PyTorch baseline for LLaMA-2 7B on an A6000 GPU \rev{(batch size 16) at prefill lengths of 2K, 8K, and 16K tokens, with and without the FlashAttention-2 (FA2) attention backend; FA2 speedups are relative to the dense model using the same backend}. Speedups decrease as the prefill grows, since attention computation begins to dominate runtime and reduces the relative contribution of accelerating the FFN GEMMs. \rev{FA2 reduces the attention overhead and increases the speedup in every setting.}}
\label{table::long_context}
\small
\setlength{\tabcolsep}{5pt}
\begin{tabular}{c c c c c c c c}
\toprule
Tokens generated & Sparsity (\%) & \rev{2K} & \rev{2K (FA2)} & 8K & \rev{8K (FA2)} & 16K & \rev{16K (FA2)} \\
\midrule
\multirow{3}{*}{128}
 & \rev{25} & \rev{1.18$\times$} & \rev{1.24$\times$} & \rev{1.09$\times$} & \rev{1.11$\times$} & \rev{1.02$\times$} & \rev{1.03$\times$} \\
 & 35 & \rev{1.27$\times$} & \rev{1.33$\times$} & 1.14$\times$ & \rev{1.17$\times$} & 1.07$\times$ & \rev{1.08$\times$} \\
 & 45 & \rev{1.38$\times$} & \rev{1.45$\times$} & 1.22$\times$ & \rev{1.26$\times$} & 1.14$\times$ & \rev{1.17$\times$} \\
\midrule
\multirow{3}{*}{\rev{1024}}
 & \rev{25} & \rev{1.13$\times$} & \rev{1.21$\times$} & \rev{1.13$\times$} & \rev{1.16$\times$} & \rev{1.12$\times$} & \rev{1.14$\times$} \\
 & \rev{35} & \rev{1.18$\times$} & \rev{1.26$\times$} & \rev{1.16$\times$} & \rev{1.22$\times$} & \rev{1.15$\times$} & \rev{1.18$\times$} \\
 & \rev{45} & \rev{1.25$\times$} & \rev{1.38$\times$} & \rev{1.19$\times$} & \rev{1.29$\times$} & \rev{1.19$\times$} & \rev{1.23$\times$} \\
\bottomrule
\end{tabular}
\end{table}

\section{Generalization to other N:M sparsity patterns}
\label{app::pattern_agnostic}

\ours is fundamentally pattern-agnostic. Our experiments use the 2:4 pattern because it is currently the only semi-structured format with native Tensor Core acceleration on modern NVIDIA GPUs, but the framework extends naturally to other N:M sparsity patterns and even to a mixture of multiple patterns within the same matrix.

Concretely, the binary tile distribution (dense vs.\ 2:4) used in Section~\ref{sec:patch} is a special case of a $K$-class tile distribution where each class corresponds to a different sparsity pattern. The sampled tile mask is then combined with a learned per-class fine-grained mask via a weighted average. For instance, augmenting \ours with 1:4 and 3:4 patterns alongside the existing dense and 2:4 options yields
\begin{equation}
    \Tilde{\mM} \;=\; \Tilde{\mM}_{\text{tile}}[0]\odot \mathbf{1} \;+\; \Tilde{\mM}_{\text{tile}}[1]\odot \Tilde{\mM}_{1:4} \;+\; \Tilde{\mM}_{\text{tile}}[2]\odot \Tilde{\mM}_{2:4} \;+\; \Tilde{\mM}_{\text{tile}}[3]\odot \Tilde{\mM}_{3:4},
    \label{eq::patch_multiclass}
\end{equation}
where $\Tilde{\mM}_{\text{tile}} \in [0,1]^{(d_1/b_1) \times (d_2/b_2) \times K}$ is sampled from a $K$-class Gumbel--Softmax. Replacing $\Tilde{\mM}_{2:4}$ in Equation~\ref{eq:tile_merge} with $\Tilde{\mM}_{N:M}$ (e.g., a 4:8 mask whose internal mask space has $\binom{8}{4}$ classes) recovers a single-pattern variant for any other N:M pattern.

This formulation has two practical implications. First, increasing the number of available patterns enlarges the global sparsity range that \ours can target: e.g., adding 1:4 (75\% sparsity) extends the upper bound from 50\% to 75\%, which would be needed to reach the 80--90\% sparsity regimes targeted by some unstructured-pruning work. Second, as new structured patterns (e.g., 4:8 on next-generation accelerators) become hardware-supported, \ours can adopt them by simply enlarging the per-tile class set, without changing the rest of the optimization. We leave a full empirical study of these multi-pattern variants to future work, since each new pattern requires its own hyperparameter sweep and hardware backend support.

\begin{revblock}
\paragraph{Newer hardware.} Evaluating \ours on newer architectures, e.g., combining it with the NVFP4 Tensor Cores of Blackwell GPUs, is a promising direction. The current limitation lies in the software stack rather than in \ours itself: STOICC, the compiler backend that enables mixed dense/2:4 tile execution, does not yet support the NVFP4 data type, so we cannot report Blackwell measurements at this time. STOICC kernels for Hopper-generation GPUs are under active development, and we expect support for newer architectures, including Blackwell, to follow the same path without requiring changes to \ours. Since sparsity and quantization are orthogonal axes of compression, and Table~\ref{table::quantization} already shows \ours composing with 4-bit weight-only quantization, we expect similar relative speedups once backend support becomes available, with the gains from sparsity and reduced precision compounding. Blackwell pairs NVFP4 with a pair-wise 4:8 structured-sparsity pattern rather than element-wise 2:4; since the formulation above is pattern-agnostic, supporting it amounts to swapping the intra-tile candidate set. We leave Blackwell/NVFP4 evaluation to future work.
\end{revblock}

\stopcontents[appendix]
\section*{NeurIPS Paper Checklist}

\begin{enumerate}

\item {\bf Claims}
    \item[] Question: Do the main claims made in the abstract and introduction accurately reflect the paper's contributions and scope?
    \item[] Answer: \answerYes{}%
    \item[] Justification: The submission introduces a learnable framework for learning 2:4/Dense tile allocation in large language models.

\item {\bf Limitations}
    \item[] Question: Does the paper discuss the limitations of the work performed by the authors?
    \item[] Answer: \answerYes{}%
    \item[] Justification: Section~\ref{sec:limitations} discusses the limitations of our work.

\item {\bf Theory assumptions and proofs}
    \item[] Question: For each theoretical result, does the paper provide the full set of assumptions and a complete (and correct) proof?
    \item[] Answer: \answerNA{}%
    \item[] Justification: No theoretical proofs.

    \item {\bf Experimental result reproducibility}
    \item[] Question: Does the paper fully disclose all the information needed to reproduce the main experimental results of the paper to the extent that it affects the main claims and/or conclusions of the paper (regardless of whether the code and data are provided or not)?
    \item[] Answer: \answerYes{}%
    \item[] Justification: All details of the experiments are included in Section~\ref{sec:results}, Appendix~\ref{app::stoicc}, Appendix~\ref{app::implementation_details}, and the Reproducibility Statement.

\item {\bf Open access to data and code}
    \item[] Question: Does the paper provide open access to the data and code, with sufficient instructions to faithfully reproduce the main experimental results, as described in supplemental material?
    \item[] Answer: \answerYes{}%
    \item[] Justification: The code is available at \url{https://github.com/Paramathic/patch} and the pruned checkpoints at \url{https://huggingface.co/collections/mohammad-mozaffari/patch}; all data used is public.

\item {\bf Experimental setting/details}
    \item[] Question: Does the paper specify all the training and test details (e.g., data splits, hyperparameters, how they were chosen, type of optimizer) necessary to understand the results?
    \item[] Answer: \answerYes{}%
    \item[] Justification: All details are included in Section~\ref{sec:results}, Appendix~\ref{app::stoicc}, and Appendix~\ref{app::implementation_details}.

\item {\bf Experiment statistical significance}
    \item[] Question: Does the paper report error bars suitably and correctly defined or other appropriate information about the statistical significance of the experiments?
    \item[] Answer: \answerNo{}%
    \item[] Justification: No error bars are reported. Appendix~\ref{app::seed} shows perplexity results across different seeds.

\item {\bf Experiments compute resources}
    \item[] Question: For each experiment, does the paper provide sufficient information on the computer resources (type of compute workers, memory, time of execution) needed to reproduce the experiments?
    \item[] Answer: \answerYes{}%
    \item[] Justification: Appendix~\ref{app::implementation_details} includes the GPUs used and the GPU-hours needed for training each model.
    
\item {\bf Code of ethics}
    \item[] Question: Does the research conducted in the paper conform, in every respect, with the NeurIPS Code of Ethics \url{https://neurips.cc/public/EthicsGuidelines}?
    \item[] Answer: \answerYes{}%
    \item[] Justification: This submission follows the NeurIPS Code of Ethics.

\item {\bf Broader impacts}
    \item[] Question: Does the paper discuss both potential positive societal impacts and negative societal impacts of the work performed?
    \item[] Answer: \answerYes{}%
    \item[] Justification: The Societal Impact statement following the main text discusses both positive and negative societal impacts.
    
\item {\bf Safeguards}
    \item[] Question: Does the paper describe safeguards that have been put in place for responsible release of data or models that have a high risk for misuse (e.g., pre-trained language models, image generators, or scraped datasets)?
    \item[] Answer: \answerYes{}%
    \item[] Justification: We release only binary pruning masks, which contain no model weights: to use them, one must obtain the original base model separately and accept its license and acceptable-use policy (the LLaMA and Gemma base models are gated behind their providers' license agreements). Sparsity degrades the quality of the model, and the effect of pruning on the safeguards of the original models has not been measured in this work. We nevertheless expect the released models to pose no risks beyond those of the original models they are derived from.

\item {\bf Licenses for existing assets}
    \item[] Question: Are the creators or original owners of assets (e.g., code, data, models), used in the paper, properly credited and are the license and terms of use explicitly mentioned and properly respected?
    \item[] Answer: \answerYes{}%
    \item[] Justification: All assets are cited in the paper and used under their licenses. \emph{Models:} LLaMA-2 7B and 13B (Llama~2 Community License), LLaMA-3.1 8B (Llama~3.1 Community License), LLaMA-3.2 1B (Llama~3.2 Community License), Gemma-3 1B (Gemma Terms of Use), and Qwen-2.5 0.5B (Apache~2.0). \emph{Data:} SlimPajama (Apache~2.0), C4 (ODC-BY), WikiText-2 (CC BY-SA~3.0), ARC (CC BY-SA~4.0), MMLU (MIT), HellaSwag (MIT), PIQA (AFL~3.0), WinoGrande and OpenBookQA (Apache~2.0), and RACE (under the terms of use published by its authors). \emph{Software:} \texttt{lm-evaluation-harness} (MIT), \texttt{transformers} and \texttt{accelerate} (Apache~2.0), PyTorch (BSD-style), Triton and STOICC (MIT), FlashAttention-2 (BSD-3-Clause), and cuSPARSELt (License Agreement for NVIDIA Software Development Kits). \emph{Baselines:} the official SparseGPT and ProxSparse (Apache~2.0) and Wanda and Thanos (MIT) implementations, and the MaskLLM checkpoints released by its authors, whose code is under the NVIDIA Source Code License; our MaskLLM reproductions use our own implementation.

\item {\bf New assets}
    \item[] Question: Are new assets introduced in the paper well documented and is the documentation provided alongside the assets?
    \item[] Answer: \answerYes{}%
    \item[] Justification: We release two assets. (1) The \ours code (\url{https://github.com/Paramathic/patch}, MIT License), with the training scripts, evaluation pipeline, and usage instructions. (2) The learned sparsity masks for every \ours model and sparsity level in the paper, together with our MaskLLM reproductions (\url{https://huggingface.co/collections/mohammad-mozaffari/patch}). Each mask is released as a bit-packed binary keep/prune mask with a loading script that applies it to the original base model, and its model card documents the base model, method, target and measured sparsity, accuracy and perplexity, training hyperparameters, and usage. Because the masks are derivatives of their base models, each is distributed under its base model's license: the Llama~2, Llama~3.1, or Llama~3.2 Community License for the LLaMA models, the Gemma Terms of Use for Gemma-3, and Apache~2.0 for Qwen-2.5. No human-subject data are involved, so no consent was required.

\item {\bf Crowdsourcing and research with human subjects}
    \item[] Question: For crowdsourcing experiments and research with human subjects, does the paper include the full text of instructions given to participants and screenshots, if applicable, as well as details about compensation (if any)? 
    \item[] Answer: \answerNA{}%
    \item[] Justification: Not a crowdsourcing research experiment.

\item {\bf Institutional review board (IRB) approvals or equivalent for research with human subjects}
    \item[] Question: Does the paper describe potential risks incurred by study participants, whether such risks were disclosed to the subjects, and whether Institutional Review Board (IRB) approvals (or an equivalent approval/review based on the requirements of your country or institution) were obtained?
    \item[] Answer: \answerNA{}%
    \item[] Justification: No crowdsourcing nor research with human subjects involved in this submission.

\item {\bf Declaration of LLM usage}
    \item[] Question: Does the paper describe the usage of LLMs if it is an important, original, or non-standard component of the core methods in this research? Note that if the LLM is used only for writing, editing, or formatting purposes and does \emph{not} impact the core methodology, scientific rigor, or originality of the research, declaration is not required.
    \item[] Answer: \answerNo{}%
    \item[] Justification: LLMs were only used for writing and formatting purposes.

\end{enumerate}

\end{document}